# Institutional Books 1.0:
# A 242B token dataset from Harvard Library's collections, refined for accuracy and usability

**Matteo Cargnelutti** [a]**, Catherine Brobston** [a]**, John Hess** [b]**, Jack Cushman** [b]**, Kristi Mukk** [b]**,**

**Aristana Scourtas** [b]**, Kyle Courtney** [c]**, Greg Leppert** ✉ [a]**, Amanda Watson** [d]**, Martha Whitehead** [c]**,**

**Jonathan Zittrain** [e]

[a] Institutional Data Initiative, Harvard Law School Library
[b] Library Innovation Lab, Harvard Law School Library
[c] Harvard Library
[d] Harvard Law School Library
[e] Harvard Law School, Harvard School of Engineering and Applied Sciences, Harvard Kennedy School

## Abstract

Large language models (LLMs) use data to learn about the world in order to produce meaningful correlations and predictions. As such, the nature, scale, quality, and diversity of the datasets used to train these models, or to support their work at inference time, have a direct impact on their quality. The rapid development and adoption of LLMs of varying quality has brought into focus the scarcity of publicly available, high-quality training data and revealed an urgent need to ground the stewardship of these datasets in sustainable practices with clear provenance chains. To that end, this technical report introduces Institutional Books 1.0, a large collection of public domain books originally digitized through Harvard Library's participation in the Google Books project, beginning in 2006. Working with Harvard Library, we extracted, analyzed, and processed these volumes into an extensively-documented dataset of historic texts. This analysis covers the entirety of Harvard Library's collection scanned as part of that project, originally spanning 1,075,899 volumes written in over 250 different languages for a total of approximately 250 billion tokens. As part of this initial release, the OCR-extracted text (original and post-processed) as well as the metadata (bibliographic, source, and generated) of the 983,004 volumes, or 242B tokens, identified as being in the public domain have been made available. This report describes this project's goals and methods as well as the results of the analyses we performed, all in service of making this historical collection more accessible and easier for humans and machines alike to filter, read and use.

✉ Corresponding author: gleppert@law.harvard.edu

# 1 Introduction

Large language models (LLMs) use data to learn about the world in order to produce meaningful correlations and predictions. As such, the nature, scale, quality, and diversity of the datasets used to train these models, or to support their work at inference time, have a direct impact on their quality (Kaplan et al., 2020; Muennighoff et al., 2023; Gunasekar et al., 2023). The rapid development and adoption of LLMs has brought into focus the scarcity of publicly available, high-quality data at the scale necessary for effective model training.

In addition, a body of recent work has described data practices that often ignore data provenance and licensing terms (Longpre et al., 2024; Baack et al., 2025); modify fundamental properties of the information, introducing data drift; rely on decontextualized data that affects training outputs (Albalak et al., 2024; Longpre et al., 2024b; Welbl et al., 2021); and often exclude any filtering options for particular use cases due to limited or nonexistent description (Dodge et al., 2021). Furthermore, the vast majority of training data that shapes the outputs and limitations of large language models is from the English speaking web (Liu et al., 2024c; Dodge et al., 2021; Gao et al., 2021; Penedo et al., 2024), creating a very limited context available to these models.

We have been encouraged by several recent examples of datasets published with the goal of making advances against one or more of these challenges. FineWeb, built on top of the web-based data of Common Crawl, focused on careful deduplication and description to facilitate filtering and improve training results (Penedo et al., 2024). Before that, The Pile prioritized data from diverse sources to make training gains (Gao et al., 2021). In their experiments, Gunasekar et al (2023) created a high quality, textbook-like training-set which reduced the amount of data needed to train their model. As well, Common Corpus (Langlais et al., 2025) recognizes the impact that high-quality library collections can have on model training.

The Institutional Data Initiative, established at the Harvard Law School Library, seeks to build on this work and establish a new paradigm of data publication and use. We believe collections from libraries and other knowledge institutions are well positioned to improve the training data ecosystem by diversifying its sources, improving documentation, strengthening provenance chains, and increasing accountability to original source material. All of this is made possible by centuries of careful stewardship by libraries of their collections (Padilla et al., 2023). By collaborating with knowledge institutions to publish their collections as well-documented datasets while working across the institutional and AI communities to co-develop data practices, we aim to create a healthier and more efficient foundation for model development across the commercial, academic, and public spheres.

To catalyze this process, this technical report introduces the Institutional Books corpus, beginning with nearly one million books digitized from Harvard Library's collections. In our processing and documentation of this dataset, we have taken a rigorous approach to information stewardship. This includes an expansive view of provenance that broadly addresses the steps followed to create this dataset; decisions about its structure and analysis meant to aid users in searching and filtering this data for their particular use case; and special care to account for the legibility and quality of the underlying data.

The output of this work is a roughly 250B token public domain dataset of digitized books and other bound material. We have refined the dataset through collection-level deduplication (see Section 4.6), OCR artifact and text analysis (see Section 4.7 and Section 4.8), and OCR text post-processing (see Section 4.9). While this collection spans multiple centuries, 60% of its contents were published between 1820 and 1920 (see Section 4.3). In addition, a volume-level analysis shows the inclusion of texts from over 250 languages, while a more nuanced analysis shows the presence of 379 languages when accounting for the occurrence of multiple languages within a given volume (see Section 4.4). Our experimental application of high-level topics classification shows thorough coverage of literature, law, philosophy, and science, along with 16 other topic areas (see Section 4.5). Finally, we have taken special care with respect to copyrighted materials, and will initially release only those volumes that we believe to be in the public domain, as denoted in HathiTrust's robust rights database[1] (see Section 5). This accounts for 983,004 volumes.

With the preliminary publication of this dataset, we further seek to establish a community-led process to grow, improve, and use institutional data in ways that strengthen the knowledge ecosystem and assert the importance of ongoing stewardship of training data from the originating knowledge institutions themselves. To this end, we are experimenting to find the best way to release this data in a manner that facilitates collaboration. We encourage input on this process to guide the full publication of this and future dataset dataset releases, beginning with the following decisions:

- At preliminary launch, we have published the metadata, including experimental metadata, in full for anyone to access and use.
- At preliminary launch, we have published the dataset including OCR-extracted text under a noncommercial license, and with a "click-through" that requires users to accept this license, additional terms of use, and to share basic contact information with us so that we can engage the community in its early use.
- At preliminary launch, we have chosen to postpone the release of the raw scan images, though we will share them liberally with researchers and libraries who wish to review them. While we know AI developers and researchers are eager for more raw materials, we believe this minor friction can help build the relationships and norms necessary to grow a collaborative community.

## 2 Contributions

With this technical report, we introduce this initial set of contributions:

1. A detailed breakdown of the analysis and processing work we have conducted on Harvard Library's Google Books collection, spanning 1,075,899 volumes, in order to make it easier for humans and machines alike to navigate and use.
2. A public dataset containing the 983,004 volumes in the collection for which there is no known copyright, accounting for 91.37% of the entire collection. For each volume, this dataset features: the original OCR-extracted text, a post-processed version of the OCR text (when suitable), as well as bibliographic, source, and generated metadata.
   Available at https://huggingface.co/datasets/instdin/institutional-books-1.0
3. The Python pipeline we created to analyze and process this collection.
   Available at https://github.com/instdin/institutional-books-1-pipeline
4. The text classification model we trained and used as part of our topic classification experiment.
   Available at https://huggingface.co/instdin/institutional-books-topic-classifier-bert

---

# 3 Retrieval of the source materials

## 3.1 Context

The collection consists of over one million books digitized through a collaboration between Harvard Library and Google Books, which began in 2006[2]. The first step of our process was to work with Harvard Library and Google to retrieve digitized materials for the entire collection from GRIN—the Google Return Interface hosted by Google to enable Google Books participating institutions to retrieve the digitized copies of their collections. This initial step required us to write a custom retrieval pipeline, which we intend to release as open-source software following further refinement.

## 3.2 Process

We used the GRIN API to list and retrieve data for all available volumes from Harvard's collection. Each volume is identified by a barcode, which was originally assigned by Harvard Library. For each barcode, we sought to retrieve an (encrypted) .tar.gz file, containing scan images, OCR data, as well as bibliographic and processing-related metadata. The majority of these volumes weren't readily available for download and instead required us to request their conversion into a downloadable format which, in addition to other queuing and rate-limiting constraints, considerably added to the time it took to retrieve the collection (at least 15 days).

Once retrieved, each `{barcode}.tar.gz` file was processed as follows:
- The full archive was decrypted and sent "as is" to a dedicated `raw` bucket.
- The OCR-extracted text, bibliographic metadata, and processing-related metadata were extracted from the archive and sent to a separate `primary` bucket as JSONL and CSV files.

## 3.3 Results

Through this process, we were able to collect a full archive for 1,004,977 out of the 1,075,899 volumes listed by Google's API. At the time of writing this paper, we do not have full clarity on the reasons why 70,922 of these volumes could not be retrieved. We hypothesize that at least a portion of these volumes were not scanned, but rather that their metadata had been uploaded in anticipation of scanning that had not occurred. This is a question we intend to revisit in future iterations.

# 4 Analysis and post-processing

## 4.1 Goals and scope

We see Institutional Books 1.0 as the beginning of a collaborative and iterative research process, and the work we undertook leading to its release focused on facilitating that process. To that end, the analysis and post-processing described in this section was designed with two goals in mind:
- To get a better understanding of the collection and share the insights we generated along the way. We hope these will help users make better informed decisions about their use of the resulting dataset.
- To complement or improve the collection's data, whenever relevant and possible, in order to make the resulting dataset easier to filter, read, and use.

---

[2] https://library.hds.harvard.edu/collections/digital/harvard-google-project

This work included some methods we approached as experimental as we did not always have an external basis for comparison to ground them. When leveraging those processes, we established clear benchmarks and, whenever possible, manually reviewed samples of the outputs to further confirm we were meeting those benchmarks. Only after satisfying these requirements did we decide to include these outputs in the final dataset. These outputs are labeled in the dataset's headings using the prefix `_gen`.

It is important to note that when working with historical materials, which comprise almost all of this collection, users are working with content that is reflective of its time and, therefore, sometimes problematic. Special care should be taken to ensure the use of research and training methods that account for the presence of offensive or harmful language, including racism, sexism, colonial attitudes, and other forms of discrimination. See our disclaimer at the end of this paper and in the dataset card for more details.

Unless specified otherwise, the analysis and post-processing steps described in this section were run on the entire collection of 1,075,899 volumes, of which the public domain volumes we are releasing are a subset. We chose to run our analysis on the full collection, including records for which we had metadata but no contents, to provide a complete picture to the Harvard community, which will have access to the full collection; because we believe the public domain subset is largely reflective of the whole; and because, as additional materials are scanned or enter the public domain, we hope to establish a process to expand the dataset at a regular cadence.

Whenever possible, we focused on a form of frugal computing (Vanderbauwhede, 2023), both to be mindful of the resources we used and to improve the reproducibility of our experiments.

The data that was both collected and generated for each volume as part of these experiments is being made available in the dataset released alongside this technical report (See appendix A).

## 4.2 Available text

### 4.2.1 Methodology

In order to better understand how much OCR-extracted text was readily available in the collection, we gathered the following metrics:

- *Total number of volumes without scanned pages.*
  We used the page count metric retrieved from GRIN, after confirming that it matched what could be observed in the OCR-extracted text export (or lack thereof).
- *Total number of volumes without OCR-extracted text.*
  We considered every volume with less than 100 `o200k_base` (OpenAI, 2022) tokens of OCR-extracted text to be textless. This choice was made in order to account for edge cases we observed in the review of raw scans from the collection. For example, some volumes may have a few scanned pages, but no meaningful text (example in Figure 1).

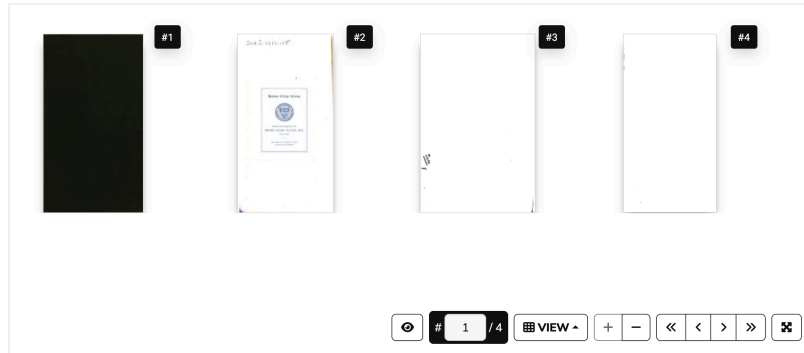

*Figure 1:* Screenshot.
Example of a volume from the collection with scanned pages, but no actual text. As seen from HathiTrust's viewer for record hvd.32044036307312.

- *Per-volume token counts.*
  We collected token counts for all available volumes using 5 different LLM tokenizers, as a way to measure the scale of the collection's OCR-extracted text. Namely, we used OpenAI's `tiktoken`[3] to run the `GPT-4` (OpenAI, 2023) and `GPT-4o` (OpenAI, 2024) tokenizers, and HuggingFace's `tokenizers` library[4] to collect token counts for `meta-llama/Llama-3.1-70B-Instruct` (Dubey et al., 2024), `microsoft/phi-4` (Abdin et al., 2024), and `mistralai/Mixtral-8x22B-Instruct-v0.1`[5]. We selected this set of target models based on their relative similarity in nature (instruct-tuned text generation models) but marked differences in number of parameters and language support.

### 4.2.2 Results

Using the metrics described in the previous subsection, we have identified that, for the 1,075,899 volumes listed in the collection:

- 394,338,216 pages were scanned, OCR-extracted, and retrievable, with an average of 367 pages per volume.
- 71,015 volumes (6.60% of the collection) do not have scanned pages. This means that, in addition to the 70,922 volumes that could not be retrieved, 93 could be retrieved but did not have any associated scanned images.
- 71,335 volumes (6.63% of the collection) have either no OCR-extracted text at all (71,094), or less than 100 `o200k_base` tokens.

By nature, token counts vary widely on a tokenizer-by-tokenizer basis. For the entirety of the OCR-extracted text of this collection, token counts range from 248B to 311B (Table 1).

---

[3] https://github.com/openai/tiktoken
[4] https://github.com/huggingface/tokenizers
[5] https://mistral.ai/news/mixtral-8x22b

*Table 1: Collection-level OCR-extracted text token counts by target LLM*

| Target LLM | Total tokens | Per-volume average |
|---|---|---|
| openai/gpt-4o | 248,299,000,580 | 230,783 |
| openai/gpt-4 | 275,637,216,999 | 256,192 |
| mistralai/Mixtral-8x22B-Instruct-v0.1 | 311,589,475,275 | 289,608 |
| microsoft/phi-4 | 275,637,216,999 | 256,192 |
| meta-llama/Llama-3.1-70B-Instruct | 267,899,787,754 | 249,001 |

Looking at the distribution of token and page counts across the collection (Figures 2 and 3) revealed that the majority of the digitized volumes are at least 100 pages long, and the majority of OCR-extracted texts are at least 100,000 `o200k_base` tokens long.

Beyond giving us insight on the availability of text at collection level, these results suggest that this collection is likely a good fit for training and evaluating models on tasks involving long context comprehension and generation, a problem domain text-generation models tend to struggle with (Li et al., 2024; Liu et al., 2024; Liu et al., 2024b).

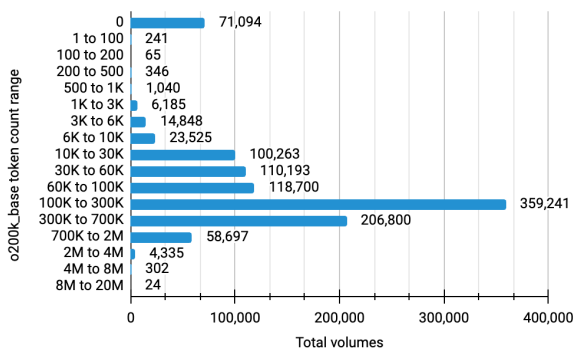

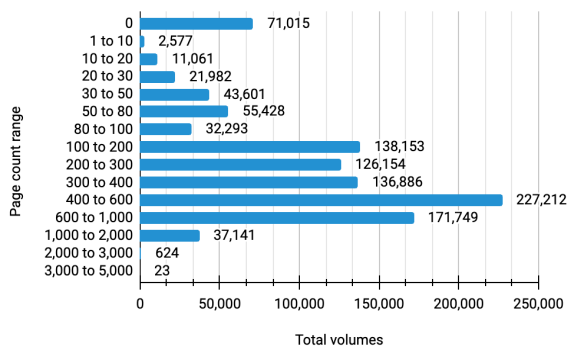

*Figure 2: Bar chart.*
Volume-level token count distribution across the collection (`o200k_base`). Notably, over 629,000 volumes contain over 100,000 tokens.

*Figure 3: Bar chart.*
Volume-level page count distribution across the collection. Over 830K volumes in the collection contain 100 pages or more.

## 4.3 Temporal coverage

### 4.3.1 Methodology

To get a coarse sense of the collection's temporal coverage, we collected and analyzed the date and date type fields from each volume's bibliographic metadata extracted from GRIN. These data points were parsed from a MARC 21 data field[6], representing bibliographic information originally provided to Google by Harvard Library. For each record, we attempted to use `Date 1` when possible, `Date 2` as a fallback, and to exclude any likely invalid date.

---

[6] https://www.loc.gov/marc/bibliographic/bd008a.html

For the purpose of this analysis we considered as invalid dates that:

- Contained a non-numeric character. For example, `u` is generally used to denote that at least part of a date is unknown (e.g: `18uu`).
- Were empty or marked as `9999`.
- Had a date type indicating that the volume is a continuing resource such as a periodical. In that case, the date fields tended to represent the publication's lifespan as opposed to a specific volume's publication date.
- Had "`No attempt to code`" in their date type.

### 4.3.2 Results

Out of 1,075,899 records, we identified that:

- 729,604 entries (67.81% of the collection) had a valid date according to the criteria outlined in the previous subsection.
- The vast majority of these dates were from the 19th and 20th centuries. 650,979 of these dates (60.55% of the collection) ranged from 1820 to 1920, with a clear spike between 1880 and 1910 (393,878 volumes, or 36.61% of the collection).

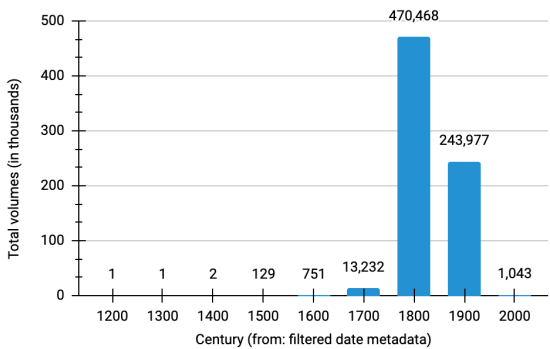
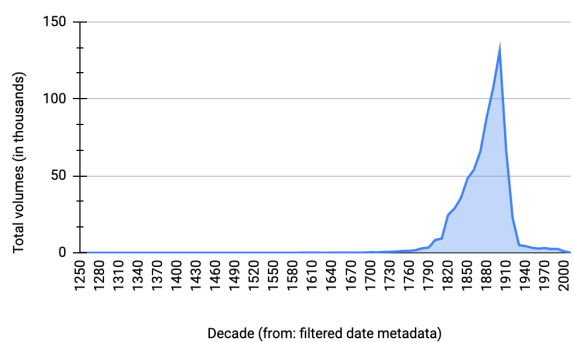

*Figure 4: Bar chart.*
Number of volumes by century, inferred from bibliographic metadata (MARC 21 date fields, filtered). Breakdown available in Appendix C.

*Figure 5: Line chart.*
Number of volumes by century, inferred from bibliographic metadata (MARC 21 date fields, filtered). Breakdown available in Appendix C.

These results suggest a high concentration of volumes from the mid-to-late 19th as well as the early 20th centuries (Figures 4 and 5). It is worth noting that 32.19% of the collection could not be dated in the context of this experiment, and that the nature of these dates varies[7]. These results therefore give us some indication, but not a complete picture, of the temporal coverage and diversity of the collection.
A detailed breakdown is available in Appendix B.

A detailed breakdown of this analysis is available in appendix B.

---

[7] https://www.loc.gov/marc/bibliographic/bd008a.html

## 4.4 Language coverage

### 4.4.1 Methodology

The multilingual capabilities of LLMs are a direct result of the language diversity of the datasets used to train them and of how well these linguistic resources are used (Gordon, Duh and Kaplan, 2021; Chang et al., 2024; Sina and Agrawal, 2024). As such, having a detailed understanding of a dataset's language coverage may prove critical as it can help facilitate decisions that result in better-performing models. To that end, we combined and compared two different language analysis methods, the results of which we incorporated in the datasets released as part of this project.

First, we parsed and analyzed the volume-level language information available in the collection's bibliographic metadata for each volume. This datapoint, originally represented as a single `ISO 639-2B` language code[8], was parsed from a MARC 21 data field representing bibliographic information originally provided by Harvard Library. We then converted this language code to `ISO 639-3` using the `iso639-lang` Python package[9].

While collecting this initial data point proved helpful, using a single language code to describe the contents of a given volume can be limiting. As a way to get a secondary signal on the "main" language of each volume and also to get a sense of the language distribution within each volume, we ran a text-detection algorithm on the OCR-extracted text of each volume.

We chose to use a Python port of the `franc` library[10] [11] on chunks of up to 768 characters, split using Langchain's `RecursiveCharacterTextSplitter`[12] . While the trigram-based detection technique used by `franc` to detect languages is simple, we found it to be well-suited to the needs of this collection:

- First and foremost because of its extensive language coverage. Indeed, the Python port of `franc` we used comes with support for 414 different languages.
- We also hypothesized that the nature of the text we needed to analyze would benefit from a conversion to trigrams in that context. Text that has been OCR-extracted from books often contains non-semantic line breaks and hyphenations, which we hypothesized that a conversion to trigram could partially mitigate.
- Finally, after initial rounds of testing and manual validation, this method appeared to be an acceptable compromise between accuracy, coverage and computational cost when accounting for the scale and scope of the collection at hand (~400M pages).

As part of this analysis, we also collected an `o200k_base` token count for each 768-character chunk we processed in order to get a language-specific token count for each volume. We also excluded from our statistics any volume-level results under 1,000 `o200k_base` tokens, which manual review revealed to be likely noise.

---

[8] https://www.loc.gov/standards/iso639-2/php/code_list.php
[9] https://github.com/LBeaudoux/iso639
[10] https://github.com/wooorm/franc
[11] https://github.com/cyb3rk0tik/pyfranc
[12] https://python.langchain.com/api_reference/text_splitters/index.html

### 4.4.2 Results

Out of 1,075,899 records, we identified:

- 241 unique volume-level languages, according to the collection's bibliographic metadata.
  72,673 volumes (6.75% of the collection) did not have volume-level language metadata.
- 254 unique volume-level languages, according to the detection we performed.
  71,656 volumes (6.66% of the collection) could not be analyzed using text-level language detection.
  This is 321 items more than the total volumes identified as missing OCR-extracted text, due to a
  variety of reasons including edge cases in the selected library and insufficient text to make a
  determination.
- 379 unique text-level languages, according to our detection.

At volume-level, the results of our detection only marginally differed from the assessment found in the
bibliographic records originally provided by Harvard Library (Figure 6 and Appendix C). This analysis
confirmed that volumes primarily written in English, other West European languages, and classical
languages such as Latin make up the vast majority of this collection. English alone represents ~47% of it.

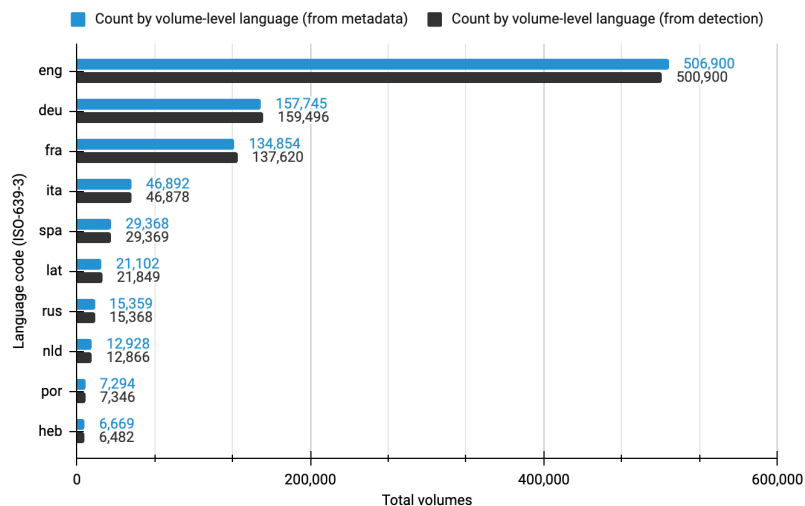

*Figure 6: Bar chart.*
Comparison of volume counts for the 10 most represented languages in the
collection, according to metadata. Blue bars represent total volumes for a given
language based on volume-level language metadata, black bars total volumes for
that same language based on our detection.

The text-level metrics we collected, although coarse, have the potential to help users make better-informed
decisions regarding data use. For example, by mapping out the language distribution within each volume
of the collection, we were able to identify books that appear to be side-by-side translations (example in
Figure 7). These texts can be particularly relevant in the context of training models for bitext mining
(Resnik, 1999) and other translations tasks.

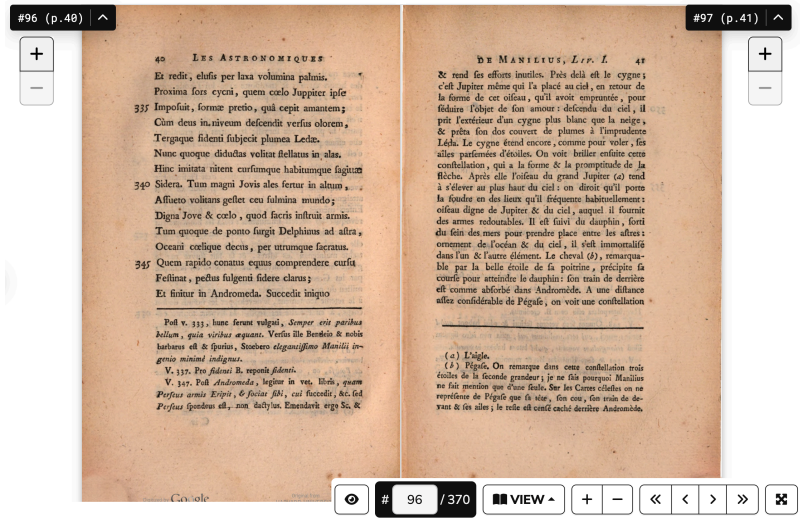

*Figure 7: Screenshot.*
Example of a volume from the collection for which text-level language detection proved useful, as it appears to be a translation which includes the original text. This volume was initially labeled as Latin; we found that it likely contains ~61% FRA tokens and ~38% LAT tokens. As seen from HathiTrust's viewer for record hvd.wl117z.

Furthermore, this text-level detection allowed us to get a sense of how much text is available in any given language, irrespective of the number of volumes using it as their "main" language (Table 2 and Appendix C). Through this lens, we observed that there are:

- 17 languages for which we detected more than 1B `o200k_base` tokens.
- 45 languages for which we detected more than 100M `o200k_base` tokens.
- 105 languages for which we detected more than 10M `o200k_base` tokens.
- And 230 languages for which we detected more than 1M `o200k_base` tokens.

*Table 2: Total detected o200k_base tokens by language code. Top 10.*

| Language code | Total detected tokens | % of total detected tokens |
|---|---|---|
| eng | 105,918,942,360 | 43.83% |
| deu | 41,803,724,013 | 17.30% |
| fra | 33,852,308,477 | 14.01% |
| ita | 9,763,407,270 | 4.04% |
| lat | 7,718,749,717 | 3.19% |
| spa | 5,424,427,269 | 2.24% |
| rus | 4,956,088,535 | 2.05% |
| ell | 3,498,189,810 | 1.45% |
| nld | 3,006,044,500 | 1.24% |
| heb | 2,376,672,753 | 0.98% |

Finally, running detection at text-level let us attempt to map-out detected languages for the 1,291 volumes originally tagged as `mul` (multiple languages) and identify a volume-level language for the 3,611 volumes originally tagged as `und` (undetermined).

The results of this analysis confirm that this collection focuses mainly on Western European languages while offering varying levels of coverage for a wide variety of languages. A few million tokens can make an important difference when training LLMs in low resource languages (Gessler and Zeldes, 2022) and these results suggest that this collection may offer meaningful support for that use case. It is worth noting that not only can detection never be fully accurate, accurate language classification sometimes requires additional context. For example, the difference between ancient and modern Greek (`grc` and `ell`) can be difficult to detect without bibliographic context. For that reason, we encourage users to compare the original language classification with the results of our detection when making decisions about their use of the collection.

Per-volume data is available in the dataset we released alongside this manuscript.

## 4.5 Topic classification

### 4.5.1 Methodology

Our goal with this experiment was to identify clear topical "tranches" within the collection. Curating and using topical datasets has proven to be an effective strategy to improve the performance of models in specialized domains and tasks, and this process generally starts with adequate classification (Parmar et al., 2024; Penedo et al., 2024).

To that end, we first collected and analyzed all of the topic/subject and form/genre information available in the collection's bibliographic metadata. We quickly identified that the majority of the records in the collection did not have a consistent topic classification but that some of this data could be used to train a classifier. We therefore filtered and used all of the topic/subject metadata we could directly map to high-level topics in order to generate a classification training set (see Appendix D).

Specifically, we used HuggingFace's `autotrain-advanced` (Thakur, 2024) to fine-tune `google-bert/bert-base-multilingual-uncased` (Devlin et al., 2019) as a text-classifier in order to assign one of 20 main classes[13] from the first level of Library of Congress' Classification Outline (LCC)[14] to each volume in the collection, using only available bibliographic metadata and the results of our language detection experiment as input signal, as illustrated in Figure 8.

---

[13] The first level of the Library of Congress' Classification Outline contains 21 items, but 2 of them are identical at that level (`E -- HISTORY OF THE AMERICAS` and `F -- HISTORY OF THE AMERICAS`).
[14] 

> **Title:** A treatise on analytical geometry of tree dimensions, containing the theory of curve surfaces and of curves of double curvature.
> **Author:** Hymers, J.
> **Year:** 1848
> **Language**: English

*Figure 8: Example.*
Example of bibliographic data presented to our topic classification model during training and inference in order to assign a high-level topic classification to individual volumes. We chose not to provide existing topic/subject and genre/form metadata as part of this prompt but to include "general note[15]" when available.

While this input format constitutes a relatively weak signal and the model we chose to train is small (168M parameters), we hypothesized that a small transformer could likely perform this task with a fairly high level of accuracy given that we had access to extensive training data for a task focused mainly on weighted word associations. We therefore chose this setup as a way to start small, improve reproducibility, and limit computational costs. In the process of assembling our training dataset, we set aside 5,000 records for validation and isolated an additional 1,000 records for benchmarking purposes to measure the accuracy of our model.

Finally, we hypothesized that the first level of the LCC would likely map well to the collection at hand. The LCC is a classification system that was first designed in the 19th century and is best suited for academic collections (Lund and Agbaji, 2018)—a description that largely matches this collection. Early experiments we conducted using the first layers of the Dewey Decimal Classification system, the Thema Category Scheme[16], and Wikipedia's topic classification system[17] revealed that the models we initially tested (text-generation models used as classifiers) performed better when using the LCC against that collection which encouraged us to continue in that direction.

### 4.5.2 Results

Out of 1,075,899 records, we first identified that:
- Only 466,356 volumes (43.35% of the collection) had any topic/subject classification metadata.
- Only 106,350 volumes (9.88% of the collection) had any form/genre classification metadata.

An analysis of the most represented values in both pre-existing classifications revealed that this data could not be used as-is to infer, at a high-level, what a given volume is about (see appendices E and F). Using the filtering mechanism described in Section 4.5.1, we identified 86,830 unique topic/subject values that we could map to the first level of the LCC, which were therefore used to build a training dataset (see Appendix G for a detailed breakdown of the training set).

---

[15] https://www.loc.gov/marc/bibliographic/bd500.html
[16] https://www.editeur.org/151/thema
[17] https://en.wikipedia.org/wiki/Category:Main_topic_classifications

After training `google-bert/bert-base-multilingual-uncased` on these examples (see training report in Appendix H), we tested it for accuracy against the classification data we set aside for benchmarking purposes (1,000 rows), on which it achieved an overall accuracy of 97.8% (Table 3).

*Table 3: Benchmarking results for our topic-classification model*
*fine-tuned from bert-base-multilingual-uncased*

| | |
|---|---|
| Total rows | 1000 |
| Matches | 978 |
| Mismatches | 22 |
| Average confidence (all) | 0.991 |
| … Standard deviation | 0.056 |
| Average confidence (mismatches) | 0.825 |
| … Standard deviation | 0.240 |
| Average confidence (matches) | 0.995 |
| … Standard deviation | 0.036 |

As part of that process, we also fine-tuned and tested `google-bert/bert-base-multilingual-cased` and `FacebookAI/xlm-roberta-large` (Conneau et al., 2020), which respectively achieved 97.1% and 95.1% accuracy against our benchmark and helped confirm our initial choice.

Finally, we ran our fine-tuned classification model against the entire collection in order to assign a high-level "topic" to each volume, resulting in the collection-level classification illustrated in Table 4.

*Table 4: Results of the topic classification experiment.*

| Topic classification | Total volumes | % of collection |
|---|---|---|
| LANGUAGE AND LITERATURE | 255,665 | 23.76% |
| LAW | 139,212 | 12.94% |
| PHILOSOPHY. PSYCHOLOGY. RELIGION | 124,617 | 11.58% |
| SCIENCE | 120,181 | 11.17% |
| SOCIAL SCIENCES | 54,865 | 5.10% |
| AGRICULTURE | 39,770 | 3.70% |
| AUXILIARY SCIENCES OF HISTORY | 36,811 | 3.42% |
| MEDICINE | 34,571 | 3.21% |
| HISTORY OF THE AMERICAS | 29,356 | 2.73% |
| POLITICAL SCIENCE | 29,279 | 2.72% |
| GEOGRAPHY. ANTHROPOLOGY. RECREATION | 25,386 | 2.36% |
| EDUCATION | 24,602 | 2.29% |
| FINE ARTS | 23,945 | 2.23% |
| TECHNOLOGY | 18,217 | 1.69% |
| MUSIC AND BOOKS ON MUSIC | 14,150 | 1.32% |
| WORLD HISTORY AND HISTORY OF EUROPE, ASIA, AFRICA, AUSTRALIA, NEW ZEALAND, ETC. | 7,839 | 0.73% |
| MILITARY SCIENCE | 7,458 | 0.69% |
| GENERAL WORKS | 7,225 | 0.67% |
| BIBLIOGRAPHY. LIBRARY SCIENCE. INFORMATION RESOURCES (GENERAL) | 5,880 | 0.55% |
| NAVAL SCIENCE | 5,482 | 0.51% |

Out of 1,075,899 records, 1,004,511 items (93.36% of the collection) could be labeled. The rest of the collection did not have sufficient metadata available, and was therefore set aside.

While not clearly indicative of the accuracy of the classification performed by the model, we collected and analyzed the confidence score of each prediction, which we averaged for each category as a way to control for the model's consistency across the classification system (Figure 9). We observed a spread ranging from 0.82 (`GENERAL WORKS`) to 0.95 (`MEDICINE`), which does not appear to directly correlate to the distribution of topics within the training set.

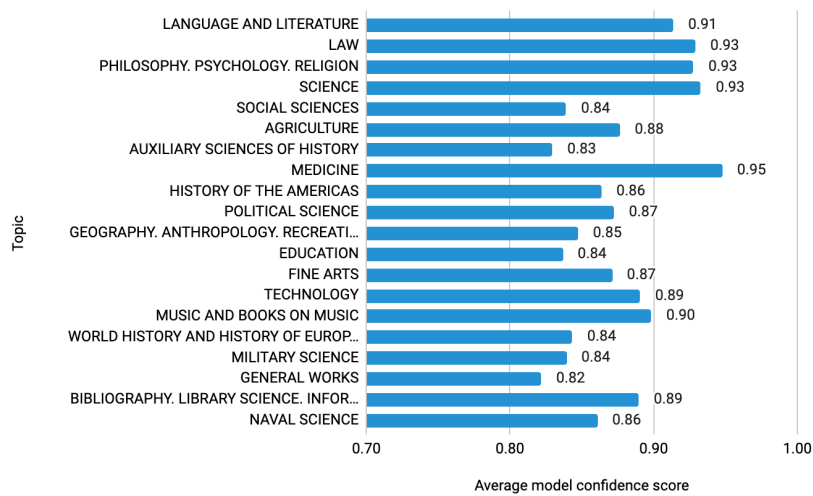

*Figure 9: Bar chart.*
Average confidence score of our fine-tuned topic classification model, by topic.
Error bars were omitted for clarity: standard deviation across data points ranges from 0.14 to 0.20, with an average of 0.19.

The results of this experiment suggest an important concentration of volumes on the topics of "`LANGUAGE AND LITERATURE`", "`LAW`", "`PHILOSOPHY. PSYCHOLOGY. RELIGION`" and "`SCIENCE`".

While this ML-assisted topic classification can be helpful, we encourage users to consider the following when using it to filter the collection:
- Benchmarking results and confidence scores are not indicative of the exactitude of the classification performed by the model. Instead, the benchmarking results reflect how well the model learned to reproduce classification patterns from the existing classification data. The confidence scores reflect how "confident" the model was in its predictions.
- The results of this topic classification experiment are intended to provide a general overview of topical coverage at collection level, and have not been reviewed by librarians at volume level.

## 4.6 Collection-level deduplication

### 4.6.1 Methodology

While research shows that deduplicating LLM pre-training data improves model performance (Lee et al., 2022; Abbas et al., 2023), there is no one-size-fits-all when it comes to deduplication strategies. Instead, they should be tailored to the nature of the underlying data and target model behaviors (Albalak et al., 2024). As such—and because we anticipate this collection to have a variety of use cases—we have chosen to focus our efforts on non-destructive, collection-level deduplication.

Our goal with this experiment was therefore to identify near-duplicate OCR-extracted texts. For that purpose, and because this deduplication process was based on text similarity, we considered different editions of the same book to be duplicates if their text was nearly identical. This necessitated finding a technique flexible enough to account for variations that result from digitizing issues, but focused enough to avoid considering two separate versions of a given book with clear additions as near-duplicates.

We first generated locality-sensitive hashes for every single OCR-extracted text in the collection. Specifically, we used a Python implementation[18] of the Simhash algorithm (Charikar, 2002) and grouped together volumes with identical hashes. Overlap in these hashes suggested the presence of near duplicate OCR-extracted texts. While often used for web content (Manku, Jain and Das Sarma, 2007), research suggests that Simhash can effectively be used on the OCR-extracted text of books (Vladimir et al., 2015).

Through trial and error, we identified that using 7-character-long shingles yielded the overall lowest level of false positives for that collection. We then used a series of heuristics to eliminate as many false positives as possible. In that context, we considered as false positives volumes in a series of detected near-duplicates that:
- Had a different volume-level detected language.
- Had a 15% or more difference in continuous character count. In that context, the continuous character count of a volume is the total number of characters in its OCR-extracted text, excluding whitespaces, line-breaks, and hyphenations. Removing these characters from the count helps to account for identical texts with slightly different layouts (e.g: the exact same text in two different print formats).

Throughout the design and implementation of this experiment, we performed accuracy control by manually reviewing a series of 100 randomly selected groups of near-duplicates. The last manual validation we performed yielded a 97% accuracy rate.

---

### 4.6.2 Results

Out of 1,075,899 records:
- We generated 1,004,681 hashes. This is more than our total of volumes with OCR-extracted text as we chose to generate simhashes for the volumes with less than 100 `o200k_base` tokens of text.
- We identified a group of 73,797 texts with at least 1 near-duplicate. Out of that group, we found 32,431 unique texts.

The above figures therefore suggest that 41,366 volumes (3.84% of the collection) are potential near-duplicates that should be set aside in certain contexts.

While we carefully designed and tested this deduplication pipeline, we chose not to use it to exclude volumes from the dataset released alongside this technical report. Instead, we elected to list the barcodes of likely near-duplicates as part of the dataset itself so users can make their own assessments in accordance with their goals.

## 4.7 OCR Artifact Analysis

### 4.7.1 Methodology

The performance of LLMs trained on OCR-extracted texts can suffer from the presence of artifacts resulting from, for example, the misinterpretation of characters (Todorov and Colavizza, 2022). Being able to measure the prevalence of these artifacts in a corpus can help inform assessments of its underlying quality for training purposes.

To perform this analysis we collected and compared two different OCR quality metrics:
- A primary volume-level OCR quality score, provided by GRIN.
- A secondary volume-level OCR quality score, which we computed by running full OCR-extracted texts against PleiAIs' `OCRoscope`[19]. This Python software library uses CLD2[20] in order to assess whether text chunks are likely to be valid text before returning an overall score (Langlais et al., 2025). The underlying technology used by this library is a limiting factor; CLD2's language support is limited to 80 languages which potentially reduces the effectiveness of `OCRoscope` on texts in languages that are not supported.

### 4.7.2 Results

Out of 1,075,899 records:
- 70,255 did not have a Google-provided OCR score. This is 667 less than the total of volumes with no archives.
- 71,260 could not be assessed using `OCROscope`. This is 338 more than the total of volumes with no archives.
- The average Google-provided OCR score is 88.38 with a standard deviation of 12.31.
- The average OCRoscope OCR score is 88.16 with a standard deviation of 16.16.

---

These collection-level averages are comparable, although higher standard deviation of OCRoscope scores suggest more variability. Plotting per-decade averages gives additional insight as to how these scores compare (Figure 10).

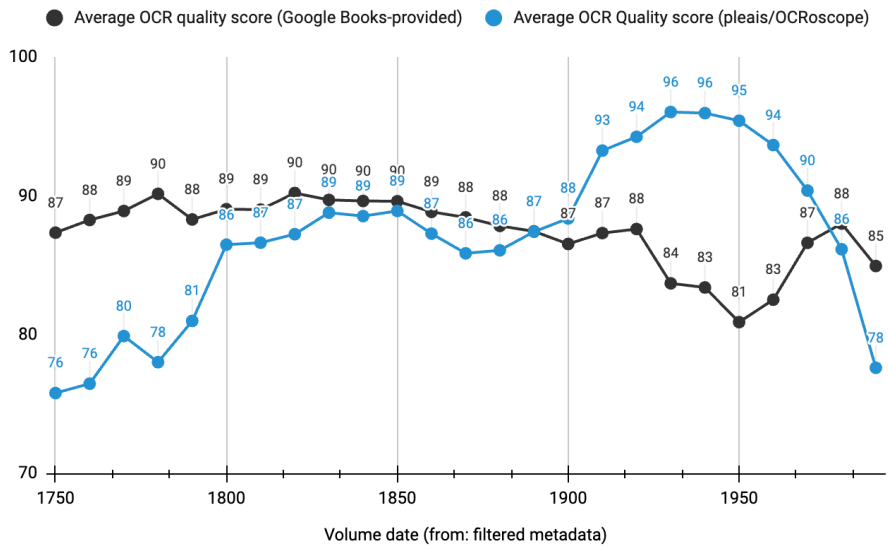

*Figure 10: Line chart.*
Comparison of Google Books and OCRoscope OCR quality score averages over time. Averages aggregated by decade. This plot focuses on the 1750 to 1990 period, which is the lengthiest continuous period of time with over 1,000 volumes per decade.
The curves indicate a 10 point difference between the two metrics for both the 18th and 20th century, which we were not able to fully interpret.
Error bars were omitted for clarity. Standard deviation for Google Books-provided scores ranged from 8.6 to 15.1 across data points, with an average of 11.35. Standard deviation for OCROscope-computed scores ranged from 8.5 to 23.14, with an average of 15.9.

These scores indicate that the collection mostly contains OCR-extracted text with limited amounts of OCR artifacts, with clear outliers. We were able to confirm this trend through the manual inspection of random samples. We acknowledge that these scores are difficult to further evaluate for accuracy without referring to the original scans, which we intend to release at a later date. Beyond collection-level analysis, we posit that combining these scores can be helpful when used at volume-level when trying to, for example, make decisions about further analysis or post-processing.

## 4.8 Text analysis

### 4.8.1 Methodology

Filtering and selecting texts suitable for the purpose of a machine learning (ML) experiment requires the use of metrics which can inform users about the underlying nature and quality of these texts.

To that end, we chose to analyze the OCR-extracted text of each volume in order to collect and share the following high-level metrics:

- *Total and unique counts at word, bigram, trigram and sentence level.*
  Text segmentation was performed using the `Polyglot` Python library[21].
- *Type-token ratios at word, bigram, trigram and sentence level.*
  Research suggests that the complexity of the texts used to train a model can affect its performance (Agrawal and Singh, 2023).
- *Average sentence length (in characters).*
  This metric may be used as a secondary text complexity or text quality metric.
- *An approximate "tokenizability" score.*
  This score, ranging from 0.0 to 100.0, indicates how efficiently `o200k_base` can encode this text. Specifically, it measures how close to 1.25 tokens per word the text is, a rough approximation of the tokenizer's average compression level
- *Character count and continuous character count.*
  For high-level filtering purposes.

These metrics were selected for their simplicity and genericity in order to help provide insight on most of the texts available in the collection, regardless of their language or nature.

### 4.8.2 Results

Combining these metrics can be helpful in filtering the collection when selecting texts in the context of an ML/NLP experiment. Furthermore, we hypothesize that some of these metrics can be used to detect the presence of scanning or OCR issues that would otherwise be difficult to identify. For example, a low "tokenizability" score for a text in a language that is well supported by `o200k_base` may suggest that the OCR-extracted text is partially unusable in its current form. Indeed, filtering records primarily written in English with a "tokenizability" score under 30.0 surfaces volumes containing mainly tables, graphs, and music sheets, which all proved challenging to capture and transcribe in the current state of this collection's OCR (example in Figures 11 and 12). Using this metric in combination with the volume-level average sentence length may further help identify such cases.

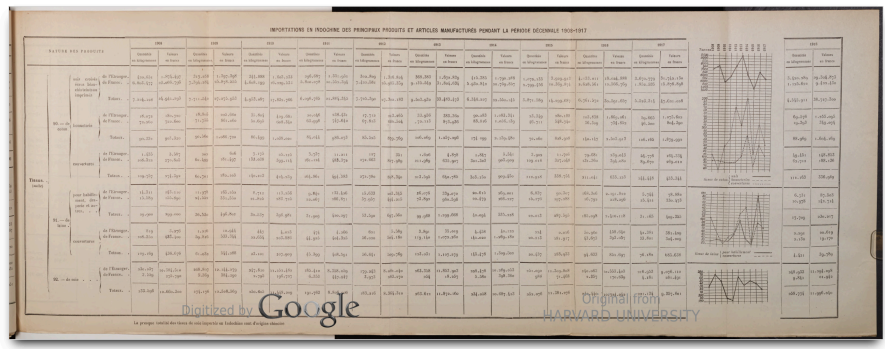

*Figure 11: Screenshot.*
Example of a document from the collection consisting mainly of tables and graphs.
The vast majority of the pages in the volume are in the format illustrated here, which the "tokenizability" score reflects.
As seen from HathiTrust's viewer for record hvd.32044088771472.

---

*Figure 12: Screenshot.*
Top, middle and end section of the resulting OCR-extracted text for the document
presented in Figure 11. This text would likely need further processing (or re-OCR)
before it could be used.

Conversely, the very nature of some of these metrics, and in particular those concerned with lexical
complexity, can be misleading when used across contexts and languages (Oh and Pellegrino, 2022). As
such we have chosen not to use them to evaluate the underlying quality of the texts at collection level.
Instead, we elected to:

- Use them as a basis of comparison for measuring the impact of post-processing (Section 4.9) and
- Provide them "as is" as part of the dataset we released.

## 4.9 OCR text post-processing

### 4.9.1 Methodology

While the word-level quality of the OCR-extracted text appears to be generally satisfactory, we observed
clear limitations. Most of these limitations are inherent to the semantic and positional decontextualization
that comes with exporting OCR data as plain text. As illustrated in figures 11 and 12, words extracted from
complex layouts such as tables, maps, illustrations, etc. are difficult to represent as plain text in a coherent
way, even when transcribed accurately. Simpler prose can also present significant challenges; in attempting
to faithfully retranscribe the contents of a scanned page, OCR pipelines generally include line breaks and
hyphenations that mirror the source content layout. This can limit the direct usability of the text in
ML/NLP contexts. More problematically, we observed that page numbers, running heads and headings
were often commingled with the rest of the text.

As a first step toward improving the usability of OCR-extracted text, we developed a post-processing
pipeline focused on addressing this category of issues. More specifically, we:

- Trained a static embedding model to detect the "type" of each OCR-extracted text line (types are
  outlined in Table 5).
- Used that coarse classification in addition to heuristics to reassemble the OCR-extracted text.

The goal of this process was to offer an alternative text output that improves machine usability for ML/NLP use cases alongside overall readability. We provide this output alongside the original OCR-extracted text exported from Google Books.

### Line-level type detection

We hypothesized that, since the OCR-extracted text of this collection is often segmented into short lines that directly map to a single type, it would be possible to get a signal on the nature of each line by:

- Using a text generation model to generate a line type detection training dataset.
- Fine-tuning a text similarity model as a classifier to perform that detection at scale.

We also posited that, since this signal would be used in combination with text-based heuristics, it did not need to be strong for the prediction to be useful. Finally, we chose to limit this post-processing experiment to the 5 most common languages in the corpus (English, German, French, Italian, and Spanish accounting for ~81% of the collection), and to only process the OCR extracted text from books with no known copyright (Section 5).

We first used `microsoft/phi-4 (14b)` (Abdin et al., 2024) to generate the training dataset for this experiment. Using the prompt described in Appendix I, we labeled individual lines from sample pages by presenting them to an 8-bit quantized version of Phi-4. For each item to annotate, we provided the model with the current, previous, and next line, as well as positional information. We used Ollama[22] to perform inference on a single A6000 GPU. The model's temperature was set to 0.0. 235,168 OCR lines from 5,000 randomly sampled pages were annotated and 10% of these samples were set aside for benchmarking purposes.

*Table 5: Generated OCR line type detection dataset, "train" split*

| Line type | Samples | % of total samples |
|---|---|---|
| PARAGRAPH_CHUNK | 94,847 | 45.01% |
| LOOSE_SENTENCE_OR_LIST_ITEM | 56,522 | 26.82% |
| NOISE_OR_BROKEN_TEXT | 27,629 | 13.11% |
| HEADING_OR_TITLE | 19,048 | 9.04% |
| PAGE_NUMBER | 7,344 | 3.49% |
| SEPARATOR | 3,682 | 1.75% |
| PARAGRAPH_END | 974 | 0.46% |
| RUNNING_HEAD | 664 | 0.32% |
| UNKNOWN | 6 | 0.00% |

The resulting training dataset was, predictably, somewhat imbalanced. While some of this imbalance is due to the nature of the documents we processed (e.g: there is likely only one running head per page), the rest is likely the result of mistakes made by the model in its classification. However, we collected enough samples for the categories we assessed to be the most critical to our core goals: assembling sentences and paragraphs, separating headings, and reducing noise (Table 5).

---

[22] https://github.com/ollama/ollama

We then used `Model2Vec`[23] to distill `sentence-transformers/LaBSE` (Feng et al., 2022; Reimers and Gurevych, 2019; Reimers and Gurevych, 2020) as a static embedding which we fine-tuned as a classifier. Distillation and fine-tuning (3 epochs) were performed on a single Apple M4 MAX SoC, taking approximately one minute each. During both fine-tuning and inference, the model was provided with very little context about the line it needed to label, limited to positional information within the page and within the volume, as illustrated in Figure 13.

```
<<12-45,5-456>> Hello world

<<{PAGE NUMBER},{TOTAL PAGES}-{LINE NUMBER}-{TOTAL-LINES}>> TEXT
```

*Figure 13: Example.*
Input format for our OCR line type detection model. Each text chunk is prefixed with positional information.

While static embedding models lack an attention mechanism (MinishLab, 2024) that would allow them to make informed use of that positional information, we reasoned that, with sufficient volume, the model could still learn patterns from this semi-structured data. The resulting model yielded a 71% accuracy rate against our benchmarking data, which matched our previously described target for coarse signal.

### *Inference and post-processing*

We then used the static embedding model we trained to help guide the post-processing of the OCR-extracted text. For each volume matching our criteria, our pipeline:

- Used the model we trained to get a signal on the "type" of each OCR line.
- Used the resulting signal as well as positional, lexical, and punctuational information to make a decision as to how the OCR line should be rendered. For example, lines detected as `HEADING_OR_TITLE` were wrapped in double line breaks, unless they were part of a series, and `PAGE_NUMBER` lines were not considered as such if they were in the middle of a page. Whenever possible, page numbers and running heads were removed.

  The details of this processing step can be found on our GitHub repository[24].

Inference and post-processing was run on a single Apple M4 MAX SoC and took approximately 5 days to complete. Inference-related statistics were collected in the process.

Finally, in order to measure the effects of these transformations on the resulting texts, we chose to perform the text analysis described in Section 4.8 on the post-processed OCR-extracted text.

### 4.9.2 Results

859,999 volumes were processed by our pipeline for a total of ~335 million pages. In the process, the static embedding model we trained performed detections on 18,405,607,403 OCR lines. Breaking these detections down by type (Table 6) confirms some of the patterns we observed during training and testing. Indeed, the model appears to detect `PARAGRAPH_CHUNK`, `LOOSE_SENTENCE_OR_LIST_ITEM`, `NOISE_OR_BROKEN_TEXT` and `HEADING_OR_TITLE` in proportions that match our overall expectations for a set of 335 million pages of text extracted mainly from books. Conversely, the model appears to largely underperform at detecting `PAGE_NUMBER` and `RUNNING_HEAD` lines, with a little over 10.5 million detections for both types combined. More surprising is the total of lines detected as `UNKNOWN`, because of how underrepresented this type was in the model's training set. Overall, these numbers matched our expectations for coarse signal, and we were able to use these OCR line type detections as a signal in our post-processing pipeline in addition to positional, lexical, and punctuational information.

---

[23] https://github.com/MinishLab/model2vec
[24] https://github.com/instdin/institutional-books-1-pipeline

*Table 6: Summary of the line-type detection process. Out of 859,999 volumes.*

| Detected line type | Number of lines | % of total detections |
|---|---|---|
| PARAGRAPH_CHUNK | 9,276,300,158 | 50.40% |
| LOOSE_SENTENCE_OR_LIST_ITEM | 4,570,972,627 | 24.83% |
| NOISE_OR_BROKEN_TEXT | 2,397,121,443 | 13.02% |
| HEADING_OR_TITLE | 1,371,820,357 | 7.45% |
| UNKNOWN | 740,229,355 | 4.02% |
| SEPARATOR | 37,576,126 | 0.20% |
| PAGE_NUMBER | 7,022,568 | 0.04% |
| RUNNING_HEAD | 4,564,506 | 0.02% |
| PARAGRAPH_END | 263 | 0.00% |

Analyzing the resulting post-processed text and comparing it to the original using the text analysis methods described in Section 4.8 revealed two distinct patterns (Figures 14, 15, 16 and 17). The first is that the average `o200k_base` "tokenizability" score (see Section 4.8.1) is consistently higher for the post-processed texts. Performing this analysis by grouping texts by language shows an average increase of 4.6 points, while doing the same comparison by grouping texts by likely decade of publication shows an average increase of 6.1 points. The second pattern is that the average detected sentence length (in characters) of the post-processed text is generally lower than that of the original OCR-extracted text. Performing this analysis by grouping texts by language shows that French is an outlier in that regard (+31 characters) while the 4 other languages we processed (English, German, Italian and Spanish) showed an average reduction of 46 characters.

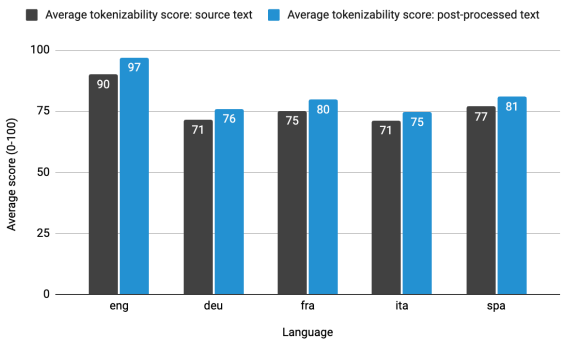

*Figure 14:* Bar plot.

Average `o200k_base` "tokenizability" score for OCR-extracted texts and their post-processed counterparts, grouped by language.

Error bars were omitted for clarity. Standard deviation across data points for the source texts ranged from 3.6 to 7.3 with an average of 4.75, and ranged from 2.8 to 4.8 with an average of 3.9 for the post-processed texts.

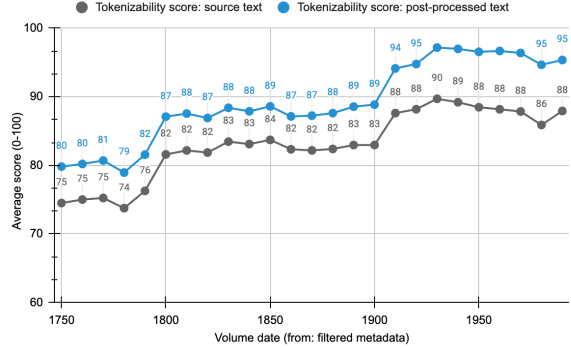

*Figure 15:* Line chart.

Average `o200k_base` "tokenizability" score for OCR-extracted texts and their post-processed counterparts, grouped by likely decade of publication. Decades included: 1750 to 1990. Error bars were omitted for clarity. Standard deviation across data points for the source texts ranged from 5.6 to 10.3 with an average of 8.9, and ranged from 5.3 to 10.9 with an average of 8.3 for the post-processed texts.

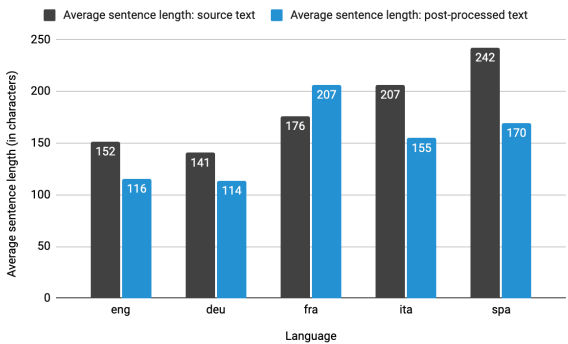

*Figure 16:* Bar plot.

Average sentence length for OCR-extracted texts and their post-processed counterparts, grouped by language. Standard deviation across data points ranged from 77 to 159 with an average of 110 for the source texts, and ranged from 53 to 87 with an average of 68 for the post-processed texts.

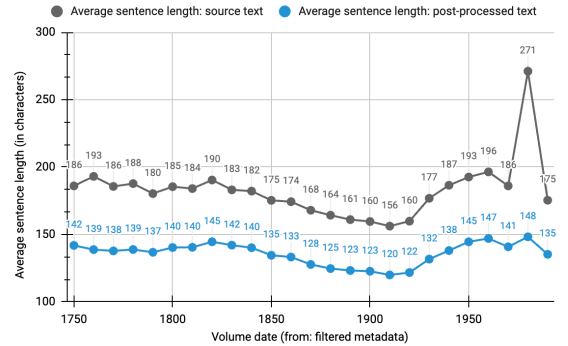

*Figure 17:* Line chart.

Average sentence length for OCR-extracted texts and their post-processed counterparts, grouped by likely decade of publication. Decades included: 1750 to 1990. Standard deviation across data points for the source texts is highly variable, ranging from 71 to 1310 with an average of 158. It ranged from 45 to 207 with an average of 76 for the post-processed texts.

We hypothesize that these metrics, while varying in accuracy, indicate that post-processing these texts made them easier to work with in certain ML/NLP contexts. Because very little text was removed in the process (-0.97% characters based on compared continuous character counts), the notable change in average sentence length and `o200k_base` "tokenizability" scores suggest a possible reduction in signals that could confuse a sentence segmenter or make a pre-trained BPE tokenizer (Sennrich, Haddow and Birch, 2016;

Gage, 1994) less efficient. In particular, the hyphenations removed in the process of reassembling paragraphs, as well as the partial removal of page numbers and running heads, may have helped in that regard. The very high variations in detected average sentence lengths in the original OCR-extracted text compared to its post-processed counterpart (figure 17) might be an additional clue pointing in that direction. For example, the text analysis data of the source texts shows that 3,001 volumes, primarily written in English, have an average detected sentence length of at least 500 characters. That figure decreased to 480 for the post-processed text. Further analysis is needed to confirm this trend and we encourage users to compare all available metrics, including their own, when making decisions about their use of this collection.

We chose to provide both the original and post-processed OCR-extracted texts as part of the dataset released alongside this technical report, as well as the text analysis metrics for both versions, so that:

- This post-processing, which is minimally but partially destructive, doesn't replace the original, full OCR-extracted text.
- Users can make a decision on which source to use based on their needs and own evaluation.

# 5 Rights determination

## 5.1 Methodology

In order to determine the copyright status of the volumes present in the collection, we used the API provided by the HathiTrust Digital Library[25] to match records for individual volumes and retrieve their current rights determination status, as inferred through their right clearances processes. Because HathiTrust preserves a copy of Harvard Library's Google Books collection and uses the collection's original barcodes as part of their `htid` identifier, matching these records proved as trivial as prefixing each barcode with a collection code (`hvd`).

## 5.2 Results

Out of 1,075,899 records, we were able to retrieve rights determination data for 1,004,497 volumes (93.36% of the collection) from HathiTrust's API. 983,510 of these volumes (91,41% of the collection) had one of the following status: `pd`, `pdus` or `cc-zero` (Table 7, full breakdown in Appendix J).

*Table 7: HathiTrust rights status distribution across the collection.*

| HathiTrust status | Total volumes | % of collection |
| --- | --- | --- |
| PD, PDUS, CC-ZERO | 983,510 | 91.41% |
| Unknown status | 75,353 | 7.00% |
| Known copyright | 16,902 | 1.57% |
| Other statuses | 134 | 0.01% |

The information retrieved through this process was used to help determine what part of the collection has no known copyright, and could therefore be included in the dataset released along this technical report. After filtering out volumes with no text (Section 4.2), 983,004 volumes were included in the dataset, for a total of 242B `o200k_base` tokens.

---

## 5.3 Rights determination statement

We respect the intellectual property rights of authors, publishers, and other rights holders. While we have taken deliberate steps to include only those volumes for which there is no known copyright restriction, specifically those identified by the HathiTrust Digital Library with a status of "public domain," "public domain in the United States," or "CC-Zero," copyright determinations are complex and context-dependent, and occasionally subject to error.

While this is relatively low risk, some volumes in this dataset may be in the public domain in the United States but still subject to copyright or other rights protections in other jurisdictions. Additionally, the absence of an explicit copyright claim or rights status does not guarantee that a work is in the public domain, either in the U.S. or abroad. Information about the copyright status of individual volumes is provided on a good-faith basis and reflects available data at the time of determination, but we cannot guarantee its completeness or accuracy.

Users of this dataset will be solely responsible for making independent legal assessments about how and where they use the materials. Some uses of materials may also be restricted by trademark, privacy, publicity rights, or other such rights or restrictions. It is the user's sole responsibility to consider the possibility that such rights or restrictions may be involved and to secure any needed permissions. If any rights holder believes that a work included in this release is misidentified or improperly included, we welcome contact and will promptly review any concerns. Our goal is to provide broad public access while maintaining respect for intellectual property rights and ensuring responsible data stewardship.

## 6 Discussion and future directions

In building, documenting, and analyzing a dataset at this scale with the primary goal of gaining insight and improving usability, we made several decisions that shaped our work. First, we wanted to release the dataset to the community in less time than it would take to come close to a complete analysis. We instead approached our version of the dataset as a launch pad for future iteration and use that could seed an ongoing collaborative publishing process. Second, we observed that we did not require access to extensive compute power to achieve this level of data refinement, so long as we were willing to creatively approach the data. Even as our compute capacity may expand, our mission requires us to heavily weight reproducibility and sustainability in our work. This will keep us grounded in a set of practices sometimes referred to as "frugal computing" (Vanderbauwhede, 2023), even as we will experiment with higher resource methods to better understand advances in the field. While we believe this is a worthwhile trade-off, it is an important element for understanding the processing choices we made. Third, we are making a conscious choice to prioritize accountability to and a public roadmap for the underlying source material in our analysis and outputs, as we believe this will yield the most productive data across commercial, academic, and public uses. This led to a decision to, for example, preserve potential duplicates within the collection without the capacity to manually confirm every result.

Looking beyond the publication of this dataset, we will expand our work across several key areas, only one of which is engineering-centric: data processing, collaboration across library and AI spheres, and the establishment of a structure for this work at scale.

First, we plan to conduct additional analysis of this dataset to expand its utility and refine processes for future work. This includes an export of the existing OCR as fully-structured text, building on the OCR post-processing work we outlined here, to make the dataset more usable for humans and machines alike. We think this is useful for both machine comprehension of individual documents, and also for traditional library use cases including more specific search and more legible OCR-extracted text for individual document review. In collaboration with libraries and the AI community we are also interested in exploring the potential for finer application of topic classification for this dataset and other large-scale datasets we release, building on the work we began here. We further plan to broaden our exploration of useful and responsible metadata to provide a richer understanding of both individual and collective texts, beginning with our upcoming work on a newspaper dataset with Boston Public Library. Finally, we see the potential to extract, describe, and release a dataset of images found within the raw page scans to support multimodal model training.

Second, we are in the process of establishing diverse collaborations to expand this dataset, improve our understanding of its contents, and sharpen our processing decisions in the future. Harvard Library is one of many Google Books partners, and we are in conversation with others who want to increase access to their scanned works. We hope the work initiated in the making of this dataset will be the beginning of a process that makes millions more books accessible to the public for a variety of uses. We also plan to partner with fellow AI research labs to evaluate how the dataset impacts model outputs. We will work with researchers in digital humanities to understand how this and other datasets from knowledge institutions can best support their work, both by making more cultural heritage material available for a variety of uses and also by providing high quality texts at scale to use as part of their own ML/NLP research. We will further explore ways to expand our use of librarian-generated metadata through the use of HathiTrust data and those from other organizations that make catalog records available at scale.

Finally, we envision this collaborative publishing process growing into an organic institutional commons that is cultivated by the community, incorporating improvements from the AI and research communities back into source datasets for collective benefit. Governed through the efforts of cross-disciplinary maintainers who guide standards and practices, such a commons would balance the need for large scale training data with a firm commitment to data integrity and stewardship by collecting institutions.

## Acknowledgements

The authors of this technical report would like to thank:
- Harvard Library for allowing us the opportunity to work with this unique collection.
- The team at the Library Innovation Lab at Harvard Law School Library, which initially incubated and significantly contributed to the success of this project. In particular, we would like to thank Ben Steinberg for providing the technical infrastructure that made most of these experiments possible.
- HuggingFace, and in particular Daniel Van Strien, Yacine Jernite and Clémentine Fourrier, who provided resources (compute and storage credits), time and expertise which helped with the preparation and release of this dataset.
- HathiTrust, and in particular Mike Furlough, Jennifer Vinopal, Kristina Hall, Janet Swatscheno, and Aaron Elkiss, for their guidance.
- The team and leadership at Google Books without whom this work would not be possible, and who assisted us in retrieving volumes and metadata and cleared the way for their release.

This work was supported by unrestricted gifts from Microsoft and OpenAI.

## Disclaimers

### Harmful Language and Content in this Dataset

This dataset is a collection of historical works that reflect the language, culture, and perspectives of their time. Users should be aware that some materials may contain language or portrayals that are outdated, offensive, or harmful today, such as racism, sexism, colonial attitudes, and other forms of discrimination. Some content may include inaccurate information, providing insight into historical contexts that existed at the time of writing. The text is maintained in its original form to retain contextual understanding and facilitate research efforts, but we encourage critical awareness and cultural sensitivity for the creators and/or subjects of the collection. These materials are offered as part of a historical perspective, but should not be considered a stand-alone research collection constructed to give a balanced perspective on any topic.

### Harmful Language in Bibliographic Description

Metadata for this collection may contain language that is overtly or implicitly harmful, outdated, or biased, or may by omission fail to represent important perspectives. Metadata may contain language created decades ago. It is common practice within the field of library science to reuse descriptions provided from the creator of the materials. While in some instances this allows communities and individuals to represent their materials in their own words, unexamined use of this practice may mean that racist or other offensive terminologies appear in our description. We also use national standardized terms in our work that can be outdated and harmful. Note that terminology in historical materials and in library descriptions does not always match the language we currently understand to be preferred by members of the communities depicted.

Furthermore, we acknowledge that the act of collecting materials is not always neutral, and the work of describing and classifying library materials is influenced by inherent personal, institutional, and societal biases. Outdated or offensive terminologies may be present in metadata such as subject headings, and harmful language or bias may be introduced by catalogers supplying titles and descriptions. In other cases, books themselves present racist, offensive or otherwise harmful viewpoints in titles or descriptions that are routinely transcribed by catalogers.

**Note:** Some language in this statement was adopted from Harvard Library's statement on Harmful Language in Library collections[26].

---

[26] https://library.harvard.edu/harmful-language-library-collections

# Reference list

Abbas, A., Tirumala, K., Simig, D., Ganguli, S. and Morcos, A.S. (2023). SemDeDup: Data-efficient learning at web-scale through semantic deduplication. *arXiv Preprint*. [online] doi:https://doi.org/10.48550/arxiv.2303.09540.

Abdin, M., Aneja, J., Behl, H., Bubeck, S., Eldan, R., Gunasekar, S., Harrison, M., Hewett, R.J., Javaheripi, M., Kauffmann, P., Lee, J.R., Lee, Y.T., Li, Y., Liu, W., Mendes, Nguyen, A., Price, E., Rosa, de, Saarikivi, O. and Salim, A. (2024). Phi-4 Technical Report. *arXiv Preprint*. [online] doi:https://doi.org/10.48550/arxiv.2412.08905.

Agrawal, A. and Singh, S. (2023). Corpus complexity matters in pretraining language models. In: S. Moosavi, I. Gurevych, Y. Hou, G. Kim, Y.J. Kim, T. Schuster and A. Agrawal, eds., *Proceedings of the Fourth Workshop on Simple and Efficient Natural Language Processing (SustaiNLP)*. [online] Association for Computational Linguistics, pp.257–263. doi:https://doi.org/10.18653/v1/2023.sustainlp-1.20.

Albalak, A., Elazar, Y., Xie, S.M., Longpre, S., Lambert, N., Wang, X., Muennighoff, N., Hou, B., Pan, L., Jeong, H., Raffel, C., Chang, S., Hashimoto, T. and Wang, W.Y. (2024). A Survey on Data Selection for Language Models. *arXiv Preprint*. [online] doi:https://doi.org/10.48550/arxiv.2402.16827.

Baack, S., Biderman, S., Odrozek, K., Skowron, A. and Wolf, T. (2025). Towards Best Practices for Open Datasets for LLM Training. *arXiv Preprint*. [online] doi:https://doi.org/10.48550/arXiv.2501.08365.

Chang, T.A., Arnett, C., Tu, Z. and Bergen, B. (2024). When is multilinguality a curse? Language modeling for 250 high- and low-resource languages. In: Y. Al-Onaizan, M. Bansal and Y.-N. Chen, eds., *Proceedings of the 2024 Conference on Empirical Methods in Natural Language Processing*. [online] Association for Computational Linguistics, pp.4074–4096. doi:https://doi.org/10.18653/v1/2024.emnlp-main.236.

Charikar, M.S. (2002). Similarity estimation techniques from rounding algorithms. In: *Proceedings of the Thiry-Fourth Annual ACM Symposium on Theory of Computing*. [online] Association for Computing Machinery, pp.380–388. doi:https://doi.org/10.1145/509907.509965.

Conneau, A., Khandelwal, K., Goyal, N., Chaudhary, V., Wenzek, G., Guzmán, F., Grave, E., Ott, M., Zettlemoyer, L. and Stoyanov, V. (2020). Unsupervised cross-lingual representation learning at scale. In: D. Jurafsky, J. Chai, N. Schluter and J. Tetreault, eds., *Proceedings of the 58th Annual Meeting of the Association for Computational Linguistics*. [online] Association for Computational Linguistics, pp.8440–8451. doi:https://doi.org/10.18653/v1/2020.acl-main.747.

Devlin, J., Chang, M.-W., Lee, K. and Toutanova, K. (2019). BERT: Pre-training of deep bidirectional transformers for language understanding. In: J. Burstein, C. Doran and T. Solorio, eds., *Proceedings of the 2019 Conference of the North American Chapter of the Association for Computational Linguistics: Human Language Technologies, Volume 1 (Long and Short Papers)*. [online] Association for Computational Linguistics, pp.4171–4186. doi:https://doi.org/10.18653/v1/N19-1423.

Dodge, J., Sap, M., Marasović, A., Agnew, W., Ilharco, G., Groeneveld, D., Mitchell, M. and Gardner, M. (2021). Documenting large webtext corpora: A case study on the colossal clean crawled corpus. In: M.-F.

Moens, X. Huang, L. Specia and S.W. Yih, eds., *Proceedings of the 2021 Conference on Empirical Methods in Natural Language Processing*. [online] Association for Computational Linguistics, pp.1286–1305. doi:https://doi.org/10.18653/v1/2021.emnlp-main.98.

Dubey, A., Jauhri, A., Pandey, A., Kadian, A., Al-Dahle, A., Letman, A., Mathur, A., Schelten, A., Yang, A., Fan, A., Goyal, A., Hartshorn, A., Yang, A., Mitra, A., Sravankumar, A., Korenev, A., Hinsvark, A., Rao, A., Zhang, A. and Rodriguez, A. (2024). The Llama 3 Herd of Models. *arXiv Preprint*. [online] doi:https://doi.org/10.48550/arxiv.2407.21783.

Feng, F., Yang, Y., Cer, D., Arivazhagan, N. and Wang, W. (2022). Language-agnostic BERT sentence embedding. In: S. Muresan, P. Nakov and A. Villavicencio, eds., *Proceedings of the 60th Annual Meeting of the Association for Computational Linguistics (Volume 1: Long Papers)*. [online] Association for Computational Linguistics, pp.878–891. doi:https://doi.org/10.18653/v1/2022.acl-long.62.

Gage, P. (1994). A new algorithm for data compression. *C Users J.*, 12, pp.23–38.

Gao, L., Biderman, S., Black, S., Golding, L., Hoppe, T., Foster, C., Phang, J., He, H., Anish Thite, Noa Nabeshima, Presser, S. and Leahy, C. (2021). The Pile: An 800GB Dataset of Diverse Text for Language Modeling. *arXiv Preprint*. [online] doi:https://doi.org/10.48550/arxiv.2101.00027.

Gessler, L. and Zeldes, A. (2022). MicroBERT: Effective Training of Low-resource Monolingual BERTs through Parameter Reduction and Multitask Learning. *arXiv Preprint*. [online] doi:https://doi.org/10.48550/arxiv.2212.12510.

Gordon, M.A., Duh, K. and Kaplan, J. (2021). Data and parameter scaling laws for neural machine translation. In: M.-F. Moens, X. Huang, L. Specia and S.W. Yih, eds., *Proceedings of the 2021 Conference on Empirical Methods in Natural Language Processing*. [online] Association for Computational Linguistics, pp.5915–5922. doi:https://doi.org/10.18653/v1/2021.emnlp-main.478.

Gunasekar, S., Zhang, Y., Aneja, J., Mendes, C.C.T., Del Giorno, A., Gopi, S., Javaheripi, M., Kauffmann, P., de Rosa, G., Saarikivi, O., Salim, A., Shah, S., Behl, H.S., Wang, X., Bubeck, S., Eldan, R., Kalai, A.T., Lee, Y.T. and Li, Y. (2023). Textbooks Are All You Need. *arXiv Preprint*. [online] doi:https://doi.org/10.48550/arXiv.2306.11644.

Kaplan, J., McCandlish, S., Henighan, T., Brown, T.B., Chess, B., Child, R., Gray, S., Radford, A., Wu, J. and Amodei, D. (2020). Scaling laws for neural language models. *arXiv Preprint*. [online] doi:https://doi.org/10.48550/arXiv.2001.08361.

Langlais, P.-C., Hinostroza, C.R., Nee, M., Arnett, C., Chizhov, P., Jones, E.K., Girard, I., Mach, D., Stasenko, A. and Yamshchikov, I.P. (2025). Common corpus: The largest collection of ethical data for LLM pre-training. *arXiv Preprint*. [online] doi:https://doi.org/10.48550/arXiv.2506.01732.

Lee, K., Ippolito, D., Nystrom, A., Zhang, C., Eck, D., Callison-Burch, C. and Carlini, N. (2022). Deduplicating training data makes language models better. In: S. Muresan, P. Nakov and A. Villavicencio, eds., *Proceedings of the 60th Annual Meeting of the Association for Computational Linguistics (Volume 1: Long Papers)*. [online] Association for Computational Linguistics, pp.8424–8445. doi:https://doi.org/10.18653/v1/2022.acl-long.577.

Li, T., Zhang, G., Do, Q.D., Yue, X. and Chen, W. (2024). Long-context LLMs Struggle with Long In-context Learning. *arXiv Preprint*. [online] doi:https://doi.org/10.48550/arxiv.2404.02060.

Liu, N.F., Lin, K., Hewitt, J., Paranjape, A., Bevilacqua, M., Petroni, F. and Liang, P. (2024a). Lost in the middle: How language models use long contexts. *Transactions of the Association for Computational Linguistics*, [online] 12, pp.157–173. doi:https://doi.org/10.1162/tacl_a_00638.

Liu, X., Dong, P., Hu, X. and Chu, X. (2024b). LongGenBench: Long-context generation benchmark. In: Y. Al-Onaizan, M. Bansal and Y.-N. Chen, eds., *Findings of the Association for Computational Linguistics: EMNLP 2024*. [online] Association for Computational Linguistics, pp.865–883. doi:https://doi.org/10.18653/v1/2024.findings-emnlp.48.

Liu, Y., Cao, J., Liu, C., Ding, K. and Jin, L. (2024c). Datasets for Large Language Models: A Comprehensive Survey. *arXiv Preprint*. [online] doi:https://doi.org/10.48550/arxiv.2402.18041.

Longpre, S., Mahari, R., Chen, A., Obeng-Marnu, N., Sileo, D., Brannon, W., Muennighoff, N., Khazam, N., Kabbara, J., Perisetla, K., Wu, X., Shippole, E., Bollacker, K., Wu, T., Villa, L., Pentland, S. and Hooker, S. (2024a). A large-scale audit of dataset licensing and attribution in AI. *Nature Machine Intelligence*, [online] 6(8), pp.975–987. doi:https://doi.org/10.1038/s42256-024-00878-8.

Longpre, S., Yauney, G., Reif, E., Lee, K., Roberts, A., Zoph, B., Zhou, D., Wei, J., Robinson, K., Mimno, D. and Ippolito, D. (2024b). A Pretrainer's Guide to Training Data: Measuring the Effects of Data Age, Domain Coverage, Quality, & Toxicity. In: *Proceedings of the 2024 Conference of the North American Chapter of the Association for Computational Linguistics: Human Language Technologies (Volume 1: Long Papers)*. [online] Association for Computational Linguistics. doi:https://doi.org/10.18653/v1/2024.naacl-long.179.

Lund, B.D. and Agbaji, D.A. (2018). What Scheme Do We Prefer? An Examination of Preference Between Library of Congress and Dewey Decimal Classification Among U.S.-Based Academic Library Employees. *KNOWLEDGE ORGANIZATION*, [online] 45(5), pp.380–392. doi:https://doi.org/10.5771/0943-7444-2018-5-380.

MinishLab (2024). Model2Vec Introduction blogpost. *Model2Vec: Distill a Small Fast Model from any Sentence Transformer*. Available at: https://minishlab.github.io/hf_blogpost/.

Muennighoff, N., Rush, A., Barak, B., Scao, L., Tazi, N., Piktus, A., Pyysalo, S., Wolf, T. and Raffel, C.A. (2023). Scaling Data-Constrained Language Models. In: *Advances in Neural Information Processing Systems 36 (NeurIPS 2023)*. [online] pp.50358–50376. Available at: https://proceedings.neurips.cc/paper_files/paper/2023/hash/9d89448b63ce1e2e8dc7af72c984c196-Abstract-Conference.html.

Oh, Y.M. and Pellegrino, F. (2022). Towards robust complexity indices in linguistic typology A corpus-based assessment. *Studies in Language. International Journal sponsored by the Foundation 'Foundations of Language'*, [online] 47(4). doi:https://doi.org/10.1075/sl.22034.oh.

OpenAI (2023). GPT-4 Technical Report. *arXiv Preprint*. [online] doi:https://doi.org/10.48550/arXiv.2303.08774.

OpenAI (2024). GPT-4o system card. *arXiv Preprint*. [online] doi:https://doi.org/10.48550/arXiv.2410.21276.

Padilla, T., Kettler, S., Varner, S. and Shorish, Y. (2023). Vancouver statement on collections as data. [online] doi:https://doi.org/10.5281/zenodo.8342171.

Parmar, J., Prabhumoye, S., Jennings, J., Liu, B., Jhunjhunwala, A., Wang, Z., Patwary, M., Shoeybi, M. and Catanzaro, B. (2024). Data, data everywhere: A guide for pretraining dataset construction. In: Y. Al-Onaizan, M. Bansal and Y.-N. Chen, eds., *Proceedings of the 2024 Conference on Empirical Methods in Natural Language Processing*. [online] Association for Computational Linguistics, pp.10671–10695. doi:https://doi.org/10.18653/v1/2024.emnlp-main.596.

Penedo, G., Kydlíček, H., Ben allal, L., Lozhkov, A., Mitchell, M., Raffel, C., Werra, V. and Wolf, T. (2024). The FineWeb Datasets: Decanting the Web for the Finest Text Data at Scale. *arXiv Preprint*. [online] doi:https://doi.org/10.48550/arxiv.2406.17557.

Reimers, N. and Gurevych, I. (2019). Sentence-BERT: Sentence embeddings using Siamese BERT-Networks. In: K. Inui, J. Jiang, V. Ng and X. Wan, eds., *Conference on Empirical Methods in Natural Language Processing and the 9th International Joint Conference on Natural Language Processing (EMNLP-IJCNLP)*. [online] Association for Computational Linguistics, pp.3982–3992. doi:https://doi.org/10.18653/v1/D19-1410.

Reimers, N. and Gurevych, I. (2020). Making monolingual sentence embeddings multilingual using knowledge distillation. In: B. Webber, T. Cohn, Y. He and Y. Liu, eds., *Proceedings of the 2020 Conference on Empirical Methods in Natural Language Processing (EMNLP)*. [online] Association for Computational Linguistics, pp.4512–4525. doi:https://doi.org/10.18653/v1/2020.emnlp-main.365.

Resnik, P. (1999). Mining the web for bilingual text. In: *Proceedings of the 37th Annual Meeting of the Association for Computational Linguistics*. [online] Association for Computational Linguistics, pp.527–534. doi:https://doi.org/10.3115/1034678.1034757.

Sennrich, R., Haddow, B. and Birch, A. (2016). Neural machine translation of rare words with subword units. In: K. Erk and N.A. Smith, eds., *Proceedings of the 54th Annual Meeting of the Association for Computational Linguistics (Volume 1: Long Papers)*. [online] Association for Computational Linguistics, pp.1715–1725. doi:https://doi.org/10.18653/v1/P16-1162.

Sina, B.N. and Agrawal, A. (2024). What drives performance in multilingual language models? In: Y. Scherrer, T. Jauhiainen, N. Ljubešić, M. Zampieri, P. Nakov and J. Tiedemann, eds., *Proceedings of the Eleventh Workshop on NLP for Similar Languages, Varieties, and Dialects (VarDial 2024)*. [online] Association for Computational Linguistics, pp.16–27. doi:https://doi.org/10.18653/v1/2024.vardial-1.2.

Manku, G.S., Jain, A. and Das Sarma, A. (2007). Detecting near-duplicates for web crawling. In: *Proceedings of the 16th international conference on World Wide Web*. [online] Association for Computing Machinery, pp.141–150. doi:https://doi.org/10.1145/1242572.1242592.

Thakur, A. (2024). AutoTrain: No-code training for state-of-the-art models. *arXiv Preprint*. [online] doi:https://doi.org/10.48550/arxiv.2410.15735.

Todorov, K. and Colavizza, G. (2022). An Assessment of the Impact of OCR Noise on Language Models. *arXiv Preprint*. [online] doi:https://doi.org/10.48550/arxiv.2202.00470.

Vanderbauwhede, W. (2023). Frugal Computing -- On the need for low-carbon and sustainable computing and the path towards zero-carbon computing. *arXiv Preprint*. [online] doi:https://doi.org/10.48550/arxiv.2303.06642.

Vladimir, K., Silic, M., Romic, N., Delac, G. and Srbljic, S. (2015). A preliminary study on similarity-preserving digital book identifiers. In: K. Zervanou, van Erp and B. Alex, eds., *Proceedings of the 9th SIGHUM Workshop on Language Technology for Cultural Heritage, Social Sciences, and Humanities (LaTeCH)*. [online] Association for Computational Linguistics, pp.78–83. doi:https://doi.org/10.18653/v1/W15-3712.

Welbl, J., Glaese, A., Uesato, J., Dathathri, S., Mellor, J., Hendricks, L.A., Anderson, K., Kohli, P., Coppin, B. and Huang, P.-S. (2021). Challenges in detoxifying language models. In: M.-F. Moens, X. Huang, L. Specia and S.W. Yih, eds., *Findings of the Association for Computational Linguistics: EMNLP 2021*. [online] Association for Computational Linguistics, pp.2447–2469. doi:https://doi.org/10.18653/v1/2021.findings-emnlp.210.

# Appendices

## Appendix A: Dataset fields

*Table App. A1: Field suffixes glossary.*

| Suffix | Description |
|--------|-------------|
| _src | *"From source".*<br>This field's data comes from information we gathered from the collection itself. |
| _gen | *"Generated".*<br>This field's data was generated as part of our analysis / post-processing. |
| _ext | *"External".*<br>This field's data was pulled from an external source via a records matching mechanism. |

*Table App. A2: Row-level fields list.*

| Field name | Type | Description | Section |
|------------|------|-------------|---------|
| barcode_src | String | The volume's barcode. Serves as a primary key/identifier. | 3 |
| title_src | String | Merge of all the title-related bibliographic metadata available for this volume. | 3 |
| author_src | String | Merge of all the author name-related bibliographic metadata available for this volume. | 3 |
| date1_src | String | First available date for that volume. Described in date_types_src. May contain placeholder characters. See MARC 21 specification for details. | 4.3 |
| date2_src | String | Second available date for that volume. | 4.3 |
| date_types_src | String | Describes the nature of date1_src and date2_src. See MARC 21 specification for details. | 4.3 |
| page_count_src | Int | Page count for that volume. | 4.2 |
| token_count_o200k_base_gen | Int | Total tokens for that volume's OCR-extracted text, as measured with o200k_base. | 4.2 |
| language_src | String | ISO 639-3 code for the main language of this book, as expressed in the collection's bibliographic metadata. Converted from original | 4.4 |
| | | ISO 639-2B for convenience. | |
| `language_gen` | String | ISO 693-3 code for the main language of this book, as detected by our text-level language analysis of the OCR-extracted text. | 4.4 |
| `language_distribution_gen` | Dict | Distribution of the languages detected by our text-level language analysis. Only languages for which more than 1000 `o200k_base` tokens were detected in total were kept. | 4.4 |
| `topic_or_subject_src` | String | Topic or subject information, as expressed in the collection's bibliographic metadata. Only available for (approximately) half of the collection. | 4.5 |
| `topic_or_subject_gen` | String | High-level "topic" assigned to this volume by our topic classification model. Inferred from existing metadata. One of the Library of Congress' Classification Outline first-level items. | 4.5 |
| `topic_or_subject_score_gen` | Float | Confidence score returned by our topic classification model for this specific prediction. | 4.5 |
| `genre_or_form_src` | String | Genre or form information, as expressed in the collection's bibliographic metadata. Only available for (approximately) 10% of the collection. | 4.5 |
| `general_note_src` | String | Additional notes about this specific volume in the collection's bibliographic metadata. | 3 |
| `ocr_score_src` | Int (0-100) | Primary OCR quality score, as expressed in the collection's metadata. | 4.7 |
| `ocr_score_gen` | Int (0-100) | Secondary OCR quality score, generated by using pleias/OCRoscope on the collection's OCR-extracted text. | 4.7 |
| `likely_duplicates_barcodes_gen` | List | List of barcodes for which the OCR-extracted text is highly-similar to this volume's. | 4.6 |
| `text_analysis_gen` | Dict | High-level text analysis of the OCR-extracted text, both original and post-processed. | 4.8 |
| `identifiers_src` | `Dict` | List of bibliographic identifiers, as expressed in the collection's metadata. | 3 |
| `hathitrust_data_ext` | `Dict` | Rights determination data pulled from the [Hathitrust API](#) for this volume. | 5 |
| `text_by_page_src` | `List[String]` | Original OCR-extracted text for this volume. | 4.2 |
| `text_by_page_gen` | `List[String]` | Post-processed OCR-extracted text for this volume. Available for books in the following languages: eng, deu, fra, ita, spa (~850K books). | 4.9 |

*Table App. A3: Fields nested under `Language_distribution_gen`*

| Field name | Type | Description |
|---|---|---|
| `languages` | `List[String]` | List of ISO 693-3 codes. Sorted by prevalence. |
| `proportion` | `List[Float]` | List of percentages. Sorted by prevalence. |

*Table App. A4: Shape of the dictionaries nested under both fields of `text_analysis_gen`.*

| Field name | Type | Description |
|---|---|---|
| `tokenizability_score` | `Float (0.0-100.0)` | Measure of how close to 1.25 `o200k_base` token per word this text is. |
| `char_count` | `Int` | Total characters. |
| `word_count` | `Int` | Total detected words (language-aware tokenization). |
| `word_count_unique` | `Int` | Total unique detected words. |
| `word_type_token_ratio` | `Float (0.0-100.0)` | Lexical diversity at word level. |
| `bigram_count` | `Int` | Total bigrams. |
| `bigram_count_unique` | `Int` | Total unique bigrams. |
| `bigram_type_token_ratio` | `Float (0.0-100.0)` | Lexical diversity at bigram level. |
| `trigram_count` | `Int` | Total bigrams. |
| `trigram_count_unique` | `Int` | Total unique bigrams. |
| `trigram_type_token_ratio` | `Float (0.0-100.0)` | Lexical diversity at trigram level. |
| `sentence_count` | `Int` | Total detected sentences. |
| `sentence_count_unique` | `Int` | Total unique detected sentences. |

*Note: Fields are identical for `text_by_page_src` and `text_by_page_gen`.*

*Table App. A4: Fields nested under* `identifiers_src`

| Field name | Type | Description |
|---|---|---|
| `lccn` | `List[String]` | List of Library of Congress Control Numbers, if available. |
| `isbn` | `List[String]` | List of International Standard Book Numbers, if available. |
| `ocolc` | `List[String]` | List of OCLC Control Numbers, if available. |

*Table App. A5: Fields nested under* `hathitrust_data_ext`

| Field name | Type | Description |
|---|---|---|
| `url` | `String` | Permalink to that volume on Hathitrust. |
| `rights_code` | `String` | [Hathitrust's rights determination code.](#) |
| `reason_code` | `String` | [Hathitrust's rights determination reason code.](#) |
| `last_check` | `String` | Date at which that information was pulled from the Hathitrust API. |

# Appendix B: Temporal coverage breakdown

*Table App.B1: Volumes by century (from: filtered metadata)*

| Century | Total volumes | % of collection |
|---|---|---|
| 1200 | 1 | 0.00% |
| 1300 | 1 | 0.00% |
| 1400 | 2 | 0.00% |
| 1500 | 129 | 0.01% |
| 1600 | 751 | 1.05% |
| 1700 | 13,232 | 1.23% |
| 1800 | 470,468 | 43.73% |
| 1900 | 243,977 | 22.68% |
| 2000 | 1,043 | 0.10% |

*Table App.B2: Volumes by decade (from: filtered metadata).*

| Decade | Total volumes | % of collection |
|---|---|---|
| 1250 | 1 | 0.00% |
| 1320 | 1 | 0.00% |
| 1420 | 1 | 0.00% |
| 1470 | 1 | 0.00% |
| 1500 | 4 | 0.00% |
| 1510 | 1 | 0.00% |
| 1520 | 2 | 0.00% |
| 1530 | 7 | 0.00% |
| 1540 | 6 | 0.00% |
| 1550 | 7 | 0.00% |
| 1560 | 13 | 0.00% |
| 1570 | 17 | 0.00% |
| 1580 | 22 | 0.00% |
| 1590 | 50 | 0.00% |
| 1600 | 52 | 0.00% |
| 1610 | 51 | 0.00% |
| 1620 | 53 | 0.00% |
| 1630 | 40 | 0.00% |
| 1640 | 57 | 0.01% |
| 1650 | 61 | 0.01% |
| 1660 | 97 | 0.01% |
| 1670 | 88 | 0.01% |
| 1680 | 104 | 0.01% |
| 1690 | 148 | 0.01% |
| 1700 | 348 | 0.03% |
| 1710 | 312 | 0.03% |
| 1720 | 524 | 0.05% |
| 1730 | 633 | 0.06% |
| 1740 | 828 | 0.08% |
| 1750 | 1,215 | 0.11% |
| 1760 | 1,240 | 0.12% |
| 1770 | 1,761 | 0.16% |
| 1780 | 2,947 | 0.27% |
| 1790 | 3,424 | 0.32% |
| 1800 | 8,374 | 0.78% |
| 1810 | 9,190 | 0.85% |
| 1820 | 24,633 | 2.29% |
| 1830 | 28,817 | 2.68% |
| 1840 | 35,690 | 3.32% |
| 1850 | 48,188 | 4.48% |
| 1860 | 53,994 | 5.02% |
| 1870 | 65,779 | 6.11% |
| 1880 | 88,436 | 8.22% |
| 1890 | 107,367 | 9.98% |
| 1900 | 131,338 | 12.21% |
| 1910 | 66,737 | 6.20% |
| 1920 | 22,601 | 2.10% |
| 1930 | 4,960 | 0.46% |
| 1940 | 4,349 | 0.40% |
| 1950 | 3,274 | 0.30% |
| 1960 | 2,735 | 0.25% |
| 1970 | 3,057 | 0.28% |
| 1980 | 2,433 | 0.23% |
| 1990 | 2,493 | 0.23% |
| 2000 | 895 | 0.08% |
| 2010 | 148 | 0.01% |

*Not listed: Decades with 0 volumes.*

# Appendix C: Language coverage breakdown

*Table App.C1: Volumes by languages, based on bibliographic metadata. Top 30.*

| Language | Total volumes | % of collection |
|----------|---------------|-----------------|
| eng | 506,900 | 47.11% |
| deu | 157,745 | 14.66% |
| fra | 134,854 | 12.53% |
| ita | 46,892 | 4.36% |
| spa | 29,368 | 2.73% |
| lat | 21,102 | 1.96% |
| rus | 15,359 | 1.43% |
| nld | 12,928 | 1.20% |
| por | 7,294 | 0.68% |
| heb | 6,669 | 0.62% |
| swe | 6,262 | 0.58% |
| zho | 6,086 | 0.57% |
| dan | 5,572 | 0.52% |
| hun | 4,994 | 0.46% |
| pol | 3,758 | 0.35% |
| und | 3,611 | 0.34% |
| ara | 3,337 | 0.31% |
| ces | 3,099 | 0.29% |
| ell | 2,715 | 0.25% |
| nor | 2,360 | 0.22% |
| isl | 2,340 | 0.22% |
| jpn | 1,850 | 0.17% |
| grc | 1,779 | 0.17% |
| cym | 1,678 | 0.16% |
| fin | 1,454 | 0.14% |
| mul | 1,291 | 0.12% |
| ota | 1,150 | 0.11% |
| hye | 885 | 0.08% |
| hrv | 872 | 0.08% |
| srp | 848 | 0.08% |

| Language | Total volumes | % of collection |
|---|---|---|
| eng | 500,900 | 46.56% |
| deu | 159,496 | 14.82% |
| fra | 137,620 | 12.79% |
| ita | 46,878 | 4.36% |
| spa | 29,369 | 2.73% |
| lat | 21,849 | 2.03% |
| rus | 15,368 | 1.43% |
| nld | 12,866 | 1.20% |
| dan | 8,014 | 0.74% |
| por | 7,346 | 0.68% |
| swe | 6,668 | 0.62% |
| ell | 6,510 | 0.61% |
| heb | 6,482 | 0.60% |
| cmn | 6,479 | 0.60% |
| hun | 4,976 | 0.46% |
| pol | 3,794 | 0.35% |
| arb | 3,506 | 0.33% |
| ces | 3,091 | 0.29% |
| isl | 2,475 | 0.23% |
| cym | 1,753 | 0.16% |
| jpn | 1,562 | 0.15% |
| oci | 1,432 | 0.13% |
| prs | 1,201 | 0.11% |
| fin | 1,198 | 0.11% |
| san | 1,046 | 0.10% |
| hye | 981 | 0.09% |
| ydd | 860 | 0.08% |
| bul | 825 | 0.08% |
| srp | 749 | 0.07% |
| ukr | 539 | 0.05% |

*Table App.C3: Total token counts (o200k_base) by detected language. Top 30.*
*Volume-level token counts < 1000 for a given language were discarded.*

| Language | Total detected tokens | % of total detected tokens |
|---|---|---|
| eng | 105,918,942,360 | 43.83% |
| deu | 41,803,724,013 | 17.30% |
| fra | 33,852,308,477 | 14.01% |
| ita | 9,763,407,270 | 4.04% |
| lat | 7,718,749,717 | 3.19% |
| spa | 5,424,427,269 | 2.24% |
| rus | 4,956,088,535 | 2.05% |
| ell | 3,498,189,810 | 1.45% |
| nld | 3,006,044,500 | 1.24% |
| heb | 2,376,672,753 | 0.98% |
| dan | 2,042,386,683 | 0.85% |
| cmn | 1,986,174,170 | 0.82% |
| sco | 1,817,812,828 | 0.75% |
| por | 1,412,080,527 | 0.58% |
| swe | 1,407,369,154 | 0.58% |
| hun | 1,342,999,109 | 0.56% |
| pol | 1,024,108,501 | 0.42% |
| ces | 969,876,169 | 0.40% |
| arb | 804,498,637 | 0.33% |
| jpn | 770,432,004 | 0.32% |
| bul | 768,212,627 | 0.32% |
| oci | 703,373,679 | 0.29% |
| ina | 651,697,782 | 0.27% |
| cym | 607,416,029 | 0.25% |
| pcm | 538,714,730 | 0.22% |
| mxi | 537,472,573 | 0.22% |
| und | 440,463,203 | 0.18% |
| hye | 329,697,846 | 0.14% |

# Appendix D: Source to LCC topic classification mapping

This mapping was used as part of our topic classification experiment in order to generate a training dataset. We used it to match individual items from the existing topic classification (source) with first-level items from the Library of Congress' Classification outline (target).Two typographical errors have been corrected in this version: A single instance of `"Swedigh"` instead of `"Swedish"`, and another single instance of `"literarure"` instead of `"literature"`.

*Table App. D: Topic classification mapping.*

| Target: LCC 1st level | Source: Existing classification |
|---|---|
| GENERAL WORKS | Encyclopedias and dictionaries, Newspapers, Periodicals |
| PHILOSOPHY. PSYCHOLOGY. RELIGION | Philosophy, Theology, Logic, Psychology, Aesthetics, Ethics, Mythology, Rationalism, Judaism, Islam, Theosophy, Buddhism, Christianity |
| AUXILIARY SCIENCES OF HISTORY | Archaeology, Numismatics, Heraldry, Genealogy, Biography |
| WORLD HISTORY AND HISTORY OF EUROPE, ASIA, AFRICA, AUSTRALIA, NEW ZEALAND, ETC. | World history |
| HISTORY OF THE AMERICAS | Indians of South America, Indians of North America |
| GEOGRAPHY. ANTHROPOLOGY. RECREATION | Geography, Cartography, Anthropology, Folklore, Manners and customs, Oceanography, Atlases, Mathematical geography |
| SOCIAL SCIENCES | Social sciences, Statistics, Commerce, Finance, Sociology, Socialism, Communism, Anarchism, Criminology |
| POLITICAL SCIENCE | Political science, Democracy, Local government, Municipal government, International relations, Representative government and representation |
| LAW | Law, Civil law, Criminal law, Constitutional law, Commercial law, Maritime law, Administrative law, Military law, Mining law, Corporation law, Educational law and legislation, Labor laws and legislation, Railroad law, Fishery law and legislation, Banking law, Marriage law, Liquor laws, Insurance law, Customary law, Patent laws and legislation, Building laws, Press law, Emigration and immigration law |
| EDUCATION | Education, Textbooks |
| MUSIC AND BOOKS ON MUSIC | Music, Piano music, Music theory, Musical notation, Orchestral music |
| FINE ARTS | Architecture, Sculpture, Drawing, Painting, Decorative arts |
| LANGUAGE AND LITERATURE | Philology, Classical philology, Oriental philology, Romance philology, Russian philology, Greek philology, Language and languages, English language, French language, German language, Latin language, Greek language, Hebrew language, Spanish language, Italian language, Arabic language, Sanskrit language, Chinese language, Indo-European languages, Russian language, Dutch language, Portuguese language, Swedish language, Irish language, Japanese language, Syriac language, Romance languages, Old Norse language, Literature, English literature, French literature, Italian literature, American literature, Russian literature, Spanish literature, Greek literature, Chinese literature, Latin literature, Polish literature, Comparative literature, Children's literature |
| SCIENCE | Science, Mathematics, Astronomy, Physics, Chemistry, Geology, Natural history, Biology, Botany, Zoology, Physiology, Human anatomy |
| MEDICINE | Medicine, Pathology, General Surgery, Ophthalmology, Gynecology, Obstetrics, Pediatrics, Dentistry, Dermatology, Therapeutics, Pharmacology, Pharmacy, Homeopathy |
| AGRICULTURE | Agriculture, Horticulture, Forests and forestry, Hunting |
| TECHNOLOGY | Technology, Engineering, Civil engineering, Electrical engineering, Electric engineering, Mechanical engineering, Mining engineering, Hydraulic engineering, Steam engineering, Home economics |
| MILITARY SCIENCE | Artillery, Military engineering, Infantry drill and tactics |
| NAVAL SCIENCE | Naval art and science, Naval architecture, Shipbuilding, Marine engineering |
| BIBLIOGRAPHY. LIBRARY SCIENCE. INFORMATION RESOURCES (GENERAL) | Library science, Bibliography, Paleography |

# Appendix E: Sample of the collection's pre-existing topic/subject classification.

*Table App. E: 20 most common values in the collection's pre-existing topic/subject classification.*

| Topic classification | Total books | % of the collection |
|---|---|---|
| Law reports, digests, etc | 13,906 | 1.29% |
| Law | 7,481 | 0.70% |
| Science | 6,729 | 0.63% |
| Education | 3,492 | 0.32% |
| Botany | 3,267 | 0.30% |
| Agriculture | 3,058 | 0.28% |
| Medicine | 2,936 | 0.27% |
| Theology | 2,745 | 0.26% |
| Natural history | 2,526 | 0.23% |
| English language | 2,357 | 0.22% |
| Legislation | 2,238 | 0.21% |
| World War, 1914-1918 | 2,230 | 0.21% |
| Jews | 2,107 | 0.20% |
| Plants | 1,831 | 0.17% |
| Railroads | 1,799 | 0.17% |
| Geography | 1,789 | 0.17% |
| Geology | 1,666 | 0.15% |
| Church history | 1,630 | 0.15% |
| History | 1,598 | 0.15% |
| French language | 1,595 | 0.15% |

# Appendix F: Sample of the collection's pre-existing genre/form classification.

*Table App. F: 20 most common values in the collection's pre-existing genre/form classification.*

| Genre/Form | Total books | % of the collection |
|---|---|---|
| History | 19,632 | 1.82% |
| Periodicals | 12,093 | 1.12% |
| Criticism, interpretation, etc | 3,650 | 0.34% |
| Computer network resources, Electronic journals | 3,113 | 0.29% |
| Fiction | 3,084 | 0.29% |
| History, Sources | 2,500 | 0.23% |
| Dictionaries | 2,140 | 0.20% |
| Biography | 2,101 | 0.20% |
| Early works | 1,807 | 0.17% |
| Bibliography | 1,588 | 0.15% |
| Electronic journals, Periodicals, Computer network resources | 1,419 | 0.13% |
| Biographies | 1,196 | 0.11% |
| Sermons | 1,137 | 0.11% |
| Poetry | 1,085 | 0.10% |
| Commentaries | 946 | 0.09% |
| Guidebooks | 877 | 0.08% |
| Drama | 826 | 0.08% |
| Fiction, History | 795 | 0.07% |
| Court decisions and opinions | 699 | 0.06% |
| Catalogs | 661 | 0.06% |
| Fiction, Juvenile works | 638 | 0.06% |

# Appendix G: Composition of the topic classification training sets

After initially testing a series of rebalancing mechanisms and observing worse performance, we chose not to rebalance these training sets.

*Table App. G: Composition of the topic classification model's training sets.*

| Target topic | Train rows | Test rows | Benchmarking rows |
|---|---|---|---|
| AGRICULTURE | 3,799 | 222 | 62 |
| AUXILIARY SCIENCES OF HISTORY | 1,831 | 116 | 24 |
| BIBLIOGRAPHY. LIBRARY SCIENCE. INFORMATION RESOURCES (GENERAL) | 416 | 35 | 3 |
| EDUCATION | 3,296 | 177 | 35 |
| FINE ARTS | 1,175 | 75 | 18 |
| GENERAL WORKS | 879 | 57 | 9 |
| GEOGRAPHY. ANTHROPOLOGY. RECREATION | 2,614 | 161 | 26 |
| HISTORY OF THE AMERICAS | 1,030 | 57 | 13 |
| LANGUAGE AND LITERATURE | 15,676 | 1012 | 169 |
| LAW | 11,560 | 678 | 155 |
| MEDICINE | 3,978 | 260 | 57 |
| MILITARY SCIENCE | 110 | 7 | 1 |
| MUSIC AND BOOKS ON MUSIC | 1,389 | 79 | 21 |
| NAVAL SCIENCE | 221 | 10 | 5 |
| PHILOSOPHY. PSYCHOLOGY. RELIGION | 7,666 | 488 | 99 |
| POLITICAL SCIENCE | 1,395 | 83 | 22 |
| SCIENCE | 19,139 | 1148 | 228 |
| SOCIAL SCIENCES | 3,165 | 250 | 33 |
| TECHNOLOGY | 980 | 55 | 12 |
| WORLD HISTORY AND HISTORY OF EUROPE, ASIA, AFRICA, AUSTRALIA, NEW ZEALAND, ETC. | 511 | 30 | 8 |

# Appendix H: Topic classification model training report

We fine-tuned `google-bert/bert-base-multilingual-uncased` for text classification using HuggingFace `autotrain-advanced`. The fine-tuning process, spanning 3 epochs, took less than 19 minutes to complete on 2 A6000 GPUs.

*Table App. H: Topic classification model fine-tuning validation metrics.*

| | |
|---|---|
| loss | 0.157407745718956 |
| f1_macro | 0.9613886456444749 |
| f1_micro | 0.9694 |
| f1_weighted | 0.9693030681223207 |
| precision_macro | 0.9679892485977634 |
| precision_micro | 0.9694 |
| precision_weighted | 0.9695713537396466 |
| recall_macro | 0.9560667596679707 |
| recall_micro | 0.9694 |
| recall_weighted | 0.9694 |
| accuracy | 0.9694 |

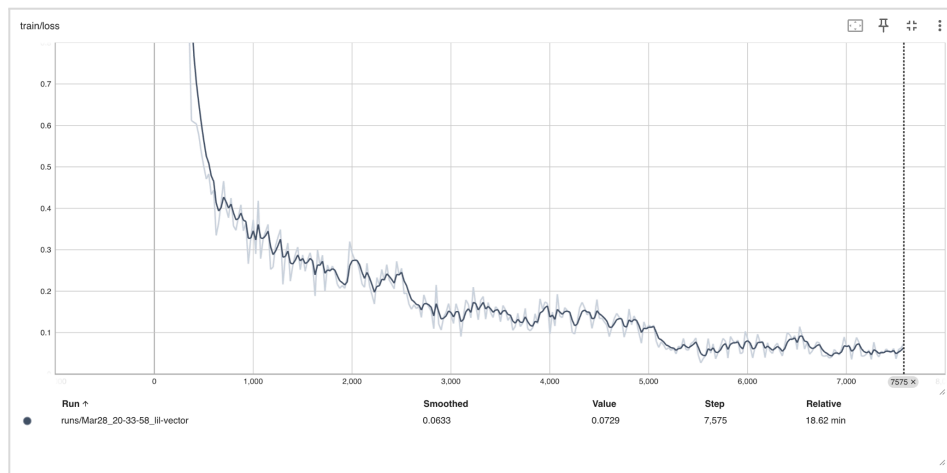

*Figure App. H: Screenshot.*
Topic classification model fine-tuning learning curve, as seen in TensorBoard's UI.

## Appendix I: OCR lines classification - Training dataset generation prompt

The initial version of this prompt included two typographical errors which have since been corrected (one missing closing > for `<context>`, as well "analyse" instead of "analyze").

```
You are a text classifier, helping with the post-processing of OCR text extracted from
books.

You will be given one or multiple text chunks to analyze as well as some contextual
information.
In this experiment, 1 chunk = 1 line extracted from a plain text OCR export.

## Information will be structured as follows:
- `<context>`: Information about the text excerpt, such as: the page number of the book this
excerpt is from, the position of this chunk on the page, and the book's main language.
- `<current>` The text chunk to analyze.
- `<previous>` The text chunk that precedes the one to analyze, if any.
- `<next>` The text chunk that follows the one to analyze, if any.

## Your role is:
- To determine the TYPE of the text chunk in `<current>`. You should use all of the
information available to help make that determination, not just the text in `<current>`.
Carefully analyze all of the information you are given.
- To return that TYPE, and nothing else. Your response MUST be one of the TYPES listed, it
cannot be anything else.

## Possible values for TYPE:
- UNKNOWN
- NOISE_OR_BROKEN_TEXT
- PAGE_NUMBER
- RUNNING_HEAD
- HEADING_OR_TITLE
- PARAGRAPH_CHUNK
- PARAGRAPH_END
- LOOSE_SENTENCE_OR_LIST_ITEM
- SEPARATOR

Carefully analyze the information you are given to accurately determine the type of the text
in `<current>`. Return a TYPE and nothing else.
```

```
<context>Page 12 of 320, Line 4 of 128</context>
<previous>Lorem ipsum </previous>
<current>dolor sit</current>
<next>amet.</next>
```

*Figure App. I:*
Prompt used to generate the OCR chunk classification dataset, alongside an example of a contextualized data chunk sent to the model alongside the prompt presented above.

# Appendix J: Hathitrust Rights Determination Breakdown

For the 1,004,497 for which rights determination data could be found on Hathitrust.

*Table App. J:  Collection-level rights determination breakdown,*
*HathiTrust data.*

| HathiTrust status | Total volumes | % of collection |
|---|---|---|
| pd | 786,964 | 73.14% |
| pdus | 196,171 | 18.23% |
| ic | 16,899 | 1.57% |
| und | 3,951 | 0.37% |
| cc-zero | 375 | 0.03% |
| cc-by-4.0 | 109 | 0.01% |
| cc-by-nc-nd-4.0 | 16 | 0.00% |
| icus | 3 | 0.00% |
| cc-by-nd-4.0 | 3 | 0.00% |
| cc-by-nc-sa-4.0 | 3 | 0.00% |
| cc-by-nc-4.0 | 3 | 0.00% |